\documentclass[10pt,twocolumn,letterpaper]{article}

\usepackage[pagenumbers]{iccv} 

\usepackage{multirow}
\usepackage{pifont}
\usepackage{algorithm}
\usepackage{algpseudocode}
\usepackage{float}
\usepackage{microtype}

\usepackage{graphicx}
\usepackage{amsmath}
\usepackage{amssymb}
\usepackage{listings}
\usepackage{xcolor}
\usepackage{enumitem}
\usepackage{tcolorbox}
\usepackage{lipsum}   

\usepackage{tabularx}        
\newcolumntype{C}{>{\centering\arraybackslash}X} 

\newcommand{\GreenCheck}{
    \textcolor{green}{\ding{51}} 
}

\tcbset{
  mybox/.style={
    colback=pink!5!white,
    colframe=pink!75!black,
    fonttitle=\bfseries,
    rounded corners,
  }
}

\definecolor{iccvblue}{rgb}{0.21,0.49,0.74}
\usepackage[pagebackref,breaklinks,colorlinks,allcolors=iccvblue]{hyperref}

\usepackage{xspace}

\newcommand{\name}{\textit{Scenethesis}\xspace}

\newcommand{\adam}{\textit{Adam} }
\newcommand{\SGD}{\textit{SGD} }

\newcommand{\rotation}{\mathbf{R}}
\newcommand{\translation}{\mathbf{T}}
\newcommand{\scale}{s}
\newcommand{\object}{o}
\newcommand{\objectref}{\tilde{o}}

\newcommand{\centroid}{\mathbf{C}}

\newcommand{\ra}[1]{\renewcommand{\arraystretch}{#1}}

\algblockdefx{Stage}{EndStage}[1]{%
  \Statex \textbf{Stage #1:}%
}{%
}

\title{\name: A Language and Vision Agentic Framework for 3D Scene Generation}

\author{Lu Ling$^{1,2}$,
Chen-Hsuan Lin$^1$,
Tsung-Yi Lin$^1$,
Yifan Ding$^1$,
Yu Zeng$^1$,
Yichen Sheng$^1$,
Yunhao Ge$^1$, \\
Ming-Yu Liu$^1$,
Aniket Bera$^{2}$\thanks{Co-last author.},
Zhaoshuo Li$^{1}$\footnotemark[1]\\
\\
$^1$NVIDIA Research \quad 
$^2$Purdue University
\\
\\
\href{https://research.nvidia.com/labs/dir/scenethesis}{https://research.nvidia.com/labs/dir/scenethesis}
}

\begin{document}

\twocolumn[
    \maketitle
    \vspace{-3em}
    \begin{center}
            \includegraphics[width=\linewidth]{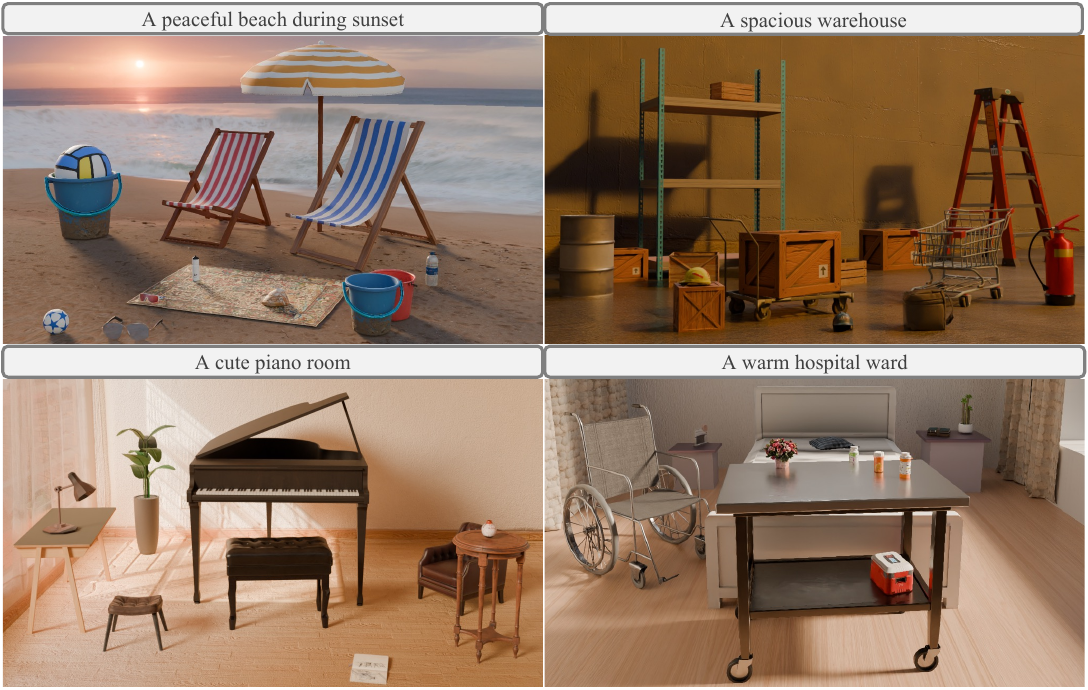}
    \captionof{figure}{\name\ is a framework for text to interactive 3D scene generation.
    Given a text prompt, \name\ leverages both language and visual priors to generate realistic and physical plausible indoor and outdoor environments.
    }
    \label{fig:teaser}

    \end{center}
]

\begingroup
\renewcommand\thefootnote{\fnsymbol{footnote}}
\footnotetext[1]{Co-last author.}
\endgroup

\begin{abstract}
Synthesizing interactive 3D scenes from text is essential for gaming, virtual reality, and embodied AI.  
However, existing methods face several challenges.
Learning-based approaches depend on small-scale indoor datasets, limiting the scene diversity and layout complexity. 
While large language models (LLMs) can leverage diverse text-domain knowledge, they struggle with spatial realism, often producing unnatural object placements that fail to respect common sense.  
Our key insight is that \textbf{vision perception can bridge this gap} by providing realistic spatial guidance that LLMs lack.  
To this end, we introduce \name, a training-free agentic framework that integrates \textbf{LLM-based scene planning with vision-guided layout refinement}.  
Given a text prompt, \name first employs an LLM to draft a coarse layout.  
A vision module then refines it by generating an image guidance and extracting scene structure to capture inter-object relations.  
Next, an optimization module iteratively enforces accurate pose alignment and physical plausibility, preventing artifacts like object penetration and instability.  
Finally, a judge module verifies spatial coherence.  
Comprehensive experiments show that \name generates diverse, realistic, and physically plausible 3D interactive scenes, making it valuable for virtual content creation, simulation environments, and embodied AI research.
\end{abstract}
   
\section{Introduction}
\label{sec:intro}
Synthesizing interactive 3D scenes from text is crucial for gaming~\cite{hu2024scenecraft}, virtual content creation~\cite{ocal2024sceneteller}, and embodied AI~\cite{yang2024holodeck,yang2024physcene,deitke2023phone2proc,kolve2017ai2, krantz2020beyond, nasiriany2024robocasa}.
Instead of generating a single scene geometry~\cite{hollein2023text2room} or differentiable rendering primitives~\cite{yu2024wonderworld}, interactive 3D scene synthesis focuses on arranging individual objects to construct a realistic layout while preserving natural interactions, function roles, and physical principles. 
For example, chairs should face tables to accommodate seating, and small items are typically placed inside cabinets, drawers, and shelves without penetration. 
Capturing these spatial relationships is crucial for generating realistic scenes, allowing virtual environments to reflect real-world structure and coherence.

Traditional interactive scene generation methods, including manual design~\cite{kolve2017ai2,gan2020threedworld, li2023behavior}, are often labor intensive and thus unscalable, while procedural approaches~\cite{deitke2022️} produce overly simplified scenes and fail to capture various real-world spatial relations. 
In recent years, deep learning-based scene generation methods, such as auto-regressive models~\cite{paschalidou2021atiss} and diffusion approaches~\cite{yang2024physcene, tang2024diffuscene}, have enabled end-to-end generation of 3D layouts. 
However, they rely on object-annotated datasets like 3D-FRONT~\cite{fu20213d}, which are small in scale, limited to indoor environments, and often contain collisions~\cite{yang2024physcene}. 
These datasets primarily model large furniture layouts while neglecting smaller objects and their functional interactions.

\begin{figure}[t]
    \centering
    \includegraphics[width=\linewidth]{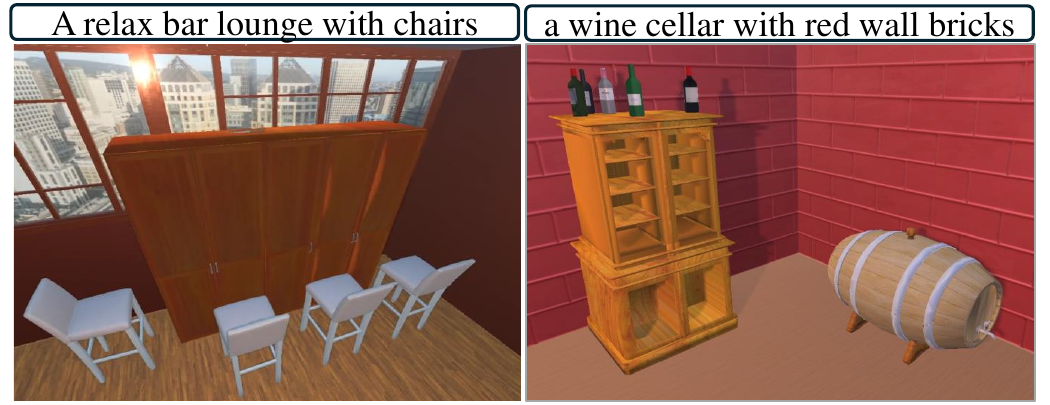}
    \caption{Unrealistic 3D scenes generated by the LLM-based method (Holodeck~\cite{yang2024holodeck}), exhibiting misplaced objects and oversimplified spatial relations.}
    \label{fig:LLM-example} \vspace{-5mm}
\end{figure}

The emergence of large language models (LLMs)~\cite{yang2024holodeck, feng2024layoutgpt, kumaran2023scenecraft} 
expands scene diversity by leveraging common-sense knowledge from text, such as which objects should co-occur based on human intent.
However, their lack of visual perception prevents them from accurately reproducing real-world spatial relations, leading to unrealistic object placements that disregard functional roles, human intent, and physical constraints.
As illustrated in~\autoref{fig:LLM-example}, LLM-generated scenes often misorient (e.g., chairs facing the cabinet) and misplace (e.g., cabinet placed against the window) objects;
small objects are restricted to predefined locations (e.g., only on top of cabinets instead of inside). 
This lack of realism disrupts object functionality,  weakens spatial coherence, and hinders structural consistency, ultimately making LLM-generated scenes impractical for real-world usability and interactions. 

Building on insights from vision foundation models that encode compact spatial information and generate coherent scene distributions reflecting real-world layouts, we introduce \name\ -- a training-free agentic framework that integrates LLM-based scene planning with vision-guided spatial refinement.
Building on top of LLMs, which lack real-world perception, \name enforces vision-based spatial constraints to enhance realism and physical plausibility.
Given a text prompt, \name employs an LLM for reasoning of coarse layout, a vision module for layout refinement, depth estimation, structural extraction, and a novel optimization for iterative alignment of object placement with visual prior through semantic correspondence matching and signed distance field (SDF)-based physical constraints, ensuring collision-free and stable integration into digital environments. 
Finally, a judge module verifies the spatial coherence.
Quantitative and qualitative results demonstrate that \name outperforms SOTA methods in scene diversity (generating indoor and outdoor scenes), layout realism, and physical plausibility. 
The layouts generated from \name can be used for downstream tasks such as virtual content creation, editing, and simulation. 
Our contribution is summarized as follows.

\begin{itemize}[]
    \item We introduce \name, a training-free agentic framework, integrates LLMs, vision foundation models, physical-aware optimization, and scene judgment to collaboratively generate realistic 3D interactive scenes.
    \item \name integrates LLM's common-sense reasoning for coarse scene planning with vision-guided spatial refinement, effectively capturing realistic inter-object relations.
    \item We propose a novel optimization process that iteratively aligns objects using semantic correspondence matching and SDF-based physical constraints, enforcing collision-free, stable, and semantically correct placements.
    \item We assess the diversity, layout realism, and object interactivity of scenes generated by \name, demonstrating superior spatial realism and physical plausibility compared to SOTA methods.
\end{itemize}

\section{Related Work}\label{sec:liter_review}

\textbf{Indoor Scene Synthesis.}  
Realistic indoor scene synthesis is essential for simulating interactive environments and training embodied agents for real-world tasks  
Early methods framed this task as layout prediction, representing scenes as graphs with object relations~\cite{luo2020end,chang2014learning,zhou2019scenegraphnet} or hierarchical structures~\cite{li2019grains,wang2019planit}.  
SceneFormer~\cite{wang2021sceneformer} and ATISS~\cite{paschalidou2021atiss} introduced autoregressive models to infer spatial relations with 3D bounding box supervision.  
Recent approaches learn layout distributions from 3D datasets like 3D-FRONT~\cite{fu20213d}, while DiffuScene~\cite{tang2024diffuscene} and InstructScene~\cite{lin2024instructscene} integrate object semantics and geometry into diffusion processes.  
PhyScene~\cite{yang2024physcene} incorporates physical constraints.  
However, interactive scene generation methods remain dataset-constrained, limiting generalization and often producing unrealistic compositions due to relaxed collision constraints~\cite{wang2021sceneformer,tang2024diffuscene}.  
Instead of learning layout distributions from limited 3D datasets, \name\ derives spatial priors from image generation models, enabling broader generalization across diverse scenarios.

\noindent\textbf{LLM/VLM Guided 3D Scene Generation.} Early efforts~\cite{deitke2022️,deitke2023phone2proc,raistrick2024infinigen} relied on rule-based procedural modeling to define spatial relations for interactive environments.  
With the rise of LLMs/VLMs, recent methods such as SceneTeller~\cite{ocal2024sceneteller}, Holodeck~\cite{yang2024holodeck}, SceneCraft~\cite{kumaran2023scenecraft}, GALA3D~\cite{zhou2024gala3d}, RobotGen~\cite{wang2023robogen}, Open-Universe~\cite{aguina2024open}, GenUSD~\cite{lin2024genusd}, LayoutVLM~\cite{sun2024layoutvlm} and SceneX~\cite{zhou2024scenex} leverage LLMs/VLMs for:
(1) spatial relation planning via predefined implicit relations,  
(2) 3D asset retrieval from semantic descriptions or vision-language embeddings, and  
(3) rule-based rough collision detection, demonstrating large-scale scene generation potential.  
Although LLMs encode rich common sense knowledge, they struggle with fine-grained spatial reasoning.  
Predefined spatial relations in text descriptions are often simplistic, limiting their ability to capture the complexity of the real-world scene~\cite{lin2024instructscene, khanna2024habitat}.  
In contrast, \name\ leverages LLM priors to convert text prompts into coarse layout instructions while using vision foundation model to persevere compact spatial information, effectively capturing real-world spatial complexity.

\noindent\textbf{Visual Foundation Model-Guided Scene Generation.}  
Visual foundation models (VFMs), particularly image generation models, have advanced visual generation and are now widely applied to 3D scene synthesis.  
Methods such as Text2Room~\cite{hollein2023text2room}, SceneScape~\cite{fridman2024scenescape}, WonderJourney~\cite{yu2024wonderjourney}, WonderWorld~\cite{yu2024wonderworld}, and Text2NeRF~\cite{zhang2024text2nerf} integrate 2D diffusion with 3D priors (e.g., depth) to generate single-geometry scenes.  
However, this approach inherently faces challenges in handling occlusions and reconstructing hidden elements due to the interconnected structure of real-world scenes, making them unsuitable for object interactions.  

Architect~\cite{wang2024architect} and Deep Prior Assembly (DPA)~\cite{zhou2024zero} introduce 2D inpainting for interactive 3D scene generation and reconstruction. While this improves occlusion handling, the lack of physical constraints and 3D reasoning leads to misaligned, floating, or intersecting objects, making it difficult to maintain functional object relationships for embodied AI tasks.
In contrast, \name\ integrates physics-aware  optimization, ensuring both spatial alignment with realistic visual prior and physical plausibility.

\noindent\textbf{Physics-Aware Scene Generation.}  
Physical principles have been largely overlooked in 3D interactive scene generation for both LLM-based and VFM-based methods. Recent works, such as PhyScene~\cite{yang2024physcene} and Holodeck~\cite{yang2024holodeck} enforce physical constraints by detecting collisions using 3D bounding boxes. 
While PhyScene reduces collision rates, it still exceeds 15\%~\cite{yang2024physcene}.  
Holodeck focuses only on large-object collision avoidance, neglecting small-object inter-collisions.  
Despite these advances, achieving full physical plausibility remains a challenge.  
To address this, \name\ incorporates precise collision detection and stability constraints, significantly reducing collision and instability rates.

\begin{figure*}[t]
    \centering
    \includegraphics[width=\linewidth]{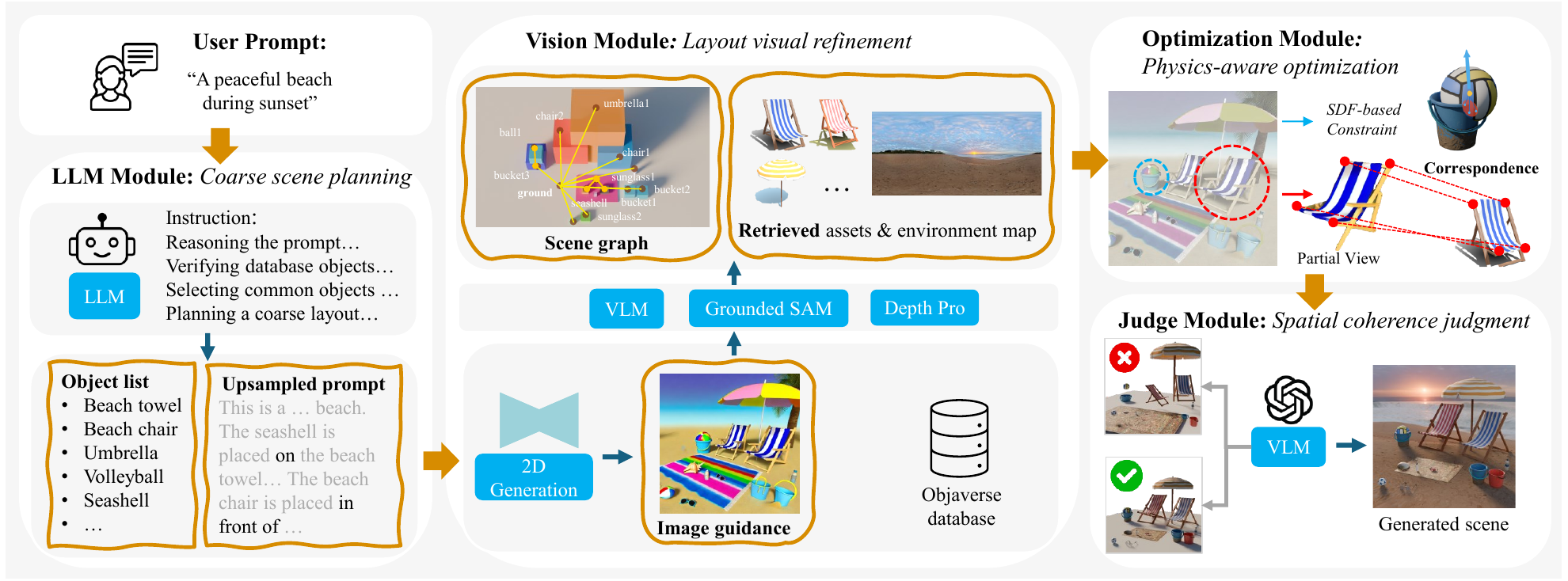}
    \captionof{figure}{\name\ is an agentic framework. The LLM module performs coarse scene planning, estimating rough spatial relationships. The vision module refines this layout by enforcing accurate spatial constraints. The physical-aware optimization iteratively adjusts object placement, ensuring pose alignment and physical plausibility. Finally, a judge module verifies the scene spatial coherence. 
    }
    \label{fig:pipeline} \vspace{-5mm}
\end{figure*}

\section{Method}
\name generates \textbf{spatially realistic, physically plausible} interactive 3D environments from user prompts.  
An overview of the pipeline is shown in~\autoref{fig:pipeline}, consisting of four key stages:  
(1) an \textbf{LLM module} drafts a coarse scene plan,  
(2) a \textbf{vision module} refines the layout with visual guidance and structural extraction,  
(3) a \textbf{physical-aware optimization module} distills priors and adjusts object placement for spatial coherence and physical plausibility, and  
(4) a \textbf{scene judge module} verifies spatial consistency.  
The following sections detail each module’s role.

\subsection{Coarse Scene Planning}
\name supports either a \textbf{simple prompt} (e.g.,~``a peaceful beach during sunset'') for flexible scene generation or a \textbf{detailed prompt} for controllable scene generation (e.g.,~a scene plan describing the detailed spatial relations as shown in the \textit{appendix}). 
For a simple prompt, the \textbf{LLM} generates a coarse scene plan by reasoning over user input. It first interprets the prompt, reviews all object categories in the available 3D database, selects commonly associated objects, and then generates an up-sampled prompt describing coarse spatial relations, as illustrated in~\autoref{fig:pipeline}.  
When given detailed prompts, the LLM checks for the presence of all specified objects in the database, infers relevant object categories, and skips the prompt up-sampling process.

Among the selected objects, the \textbf{LLM} identifies an \textit{anchor} object, following prior work~\cite{yang2024holodeck}.  
The anchor serves as the central reference point, occupying the highest spatial hierarchy apart from the \textit{ground}.  
Then the LLM establishes a coarse spatial hierarchy, positioning objects relative to the anchor and incorporating these relationships into the upsampled prompt.  
For example, in a cozy living room, the sofa acts as the anchor at the \textit{center}, while a bookshelf is placed in the \textit{background}, \textit{aligned against} the wall.  
Other objects, such as a coffee table or chairs, are positioned \textit{in front of} or \textit{beside} the sofa.

\subsection{Layout Visual Refinement}
A key insight of \name\ is that image generation models inherently encode object functionality and spatial relationships by learning common co-occurrences and spatial arrangements from large-scale image datasets.  
The \textbf{vision module} refines the coarse layout through three steps:  
(1) \textit{Image Guidance} – Generates images to refine spatial relations, ensuring realism and object functionality.  
(2) \textit{Scene Graph Generation} – Segments objects, estimates depth and 3D bounding boxes, and constructs a graph encoding inter-object relationships to establish the initial layout.  
(3) \textit{Asset Retrieval} – Selects 3D assets and environment maps for final scene composition.

\noindent\textbf{Image Generation.}  
The \textbf{vision module} refines the upsampled prompt into a visually structured scene representation. This generated image serves as the basis for segmentation, depth estimation, and asset retrieval.  

\noindent\textbf{Scene Graph Generation.}  
Leveraging vision foundation models such as GPT-4o~\cite{hurst2024gpt}, Grounded-SAM~\cite{ren2024grounded}, and DepthPro~\cite{bochkovskii2024depth}, the vision module constructs a \textbf{scene graph} that localizes objects using 3D bounding boxes (3DBB) and identifies structural components, including the \textit{anchor object}, \textit{parent objects}, and \textit{child objects} (see~\autoref{fig:pipeline}).  

To initialize asset 5DoF poses, vision module segments objects using semantic cues, estimates depth maps, and projects them into a 3D point cloud. However, due to occlusion, limited perspectives, and segmentation errors, cropped image guidance may miss full object visibility, leading to biases in 3DBB estimation -- necessitating pose adjustments later (\autoref{sssec:pose_alignment}).

The scene graph forms the basis for iterative 5DoF pose adjustments during optimization in the next stage.  
Since \name\ focuses on ground-level object layout, background elements, e.g.~wall decorations, are visually defined by the retrieved environment map.  
Detailed scene graph formatting instructions are provided in the \textit{appendix}.

\noindent\textbf{Asset Retrieval.}
Unlike existing 3D object generation and reconstruction techniques~\cite{liang2024luciddreamer, wu2024reconfusion}, such as 3D Gaussian splatting, which can produce photorealistic visuals but suffer from artifacts and geometric inconsistencies. 
These methods lack editable meshes, UV mappings, and decomposable PBR materials, making them incompatible with standard production workflows. 
To address these limitations, \name\ adopts a retrieval-based approach for asset selection, ensuring both geometric fidelity and editability for downstream applications.  
We construct a high-quality asset subset from Objaverse~\cite{deitke2023objaverse} similar to Holodeck~\cite{yang2024holodeck}, and supplemented with a custom environment map dataset.  
In the final step, the 3D assets and an environment map are retrieved to assemble a visually coherent scene.  
Retrieval details can be found in \textit{appendix}.

\subsection{Physics-aware Optimization}

Directly placing 3D assets based on estimated point clouds from image guidance poses significant challenges:  
(1) \emph{Occlusions} in real-world scenarios result in incomplete 3D point clouds, leading to errors in object orientation, scale, and position.  
(2) \emph{Discrepancies} between retrieved assets and image guidance in texture and shape make precise pose estimation difficult.
To overcome these issues, \name\ employs a \textbf{physics-aware optimization} powered by robust semantic feature matching~\cite{zhang2024telling,edstedt2024roma,chen2024secondpose} and signed-distance fields (SDFs).  
This optimization process iteratively refines object poses to ensure pose alignment and physical plausibility.

\subsubsection{Pose Alignment}
\label{sssec:pose_alignment}

To address pose estimation errors from occlusions, segmentation, or asset mismatches, we adopt dense correspondence matching from RoMa~\cite{edstedt2024roma}, leveraging semantic spatial features for robustness to occlusions and partial views.  
Unavoidable discrepancies in texture and shape between image guidance and retrieved assets are mitigated by focusing on high-level semantics over low-level details.  

For each object, we match $N$ correspondences between the rendered object and partially visible regions in the image guidance in 2D space.
It then minimizes MSE loss on both 2D and 3D spatial locations of these $N$ correspondences, backpropagating gradients to refine scale, translation, and upright rotation, as shown in~\autoref{fig:pipeline}. 
Further details on pose estimation are provided in the \textit{Appendix}.

\begin{figure}[t]
    \centering
    \includegraphics[width=\linewidth]{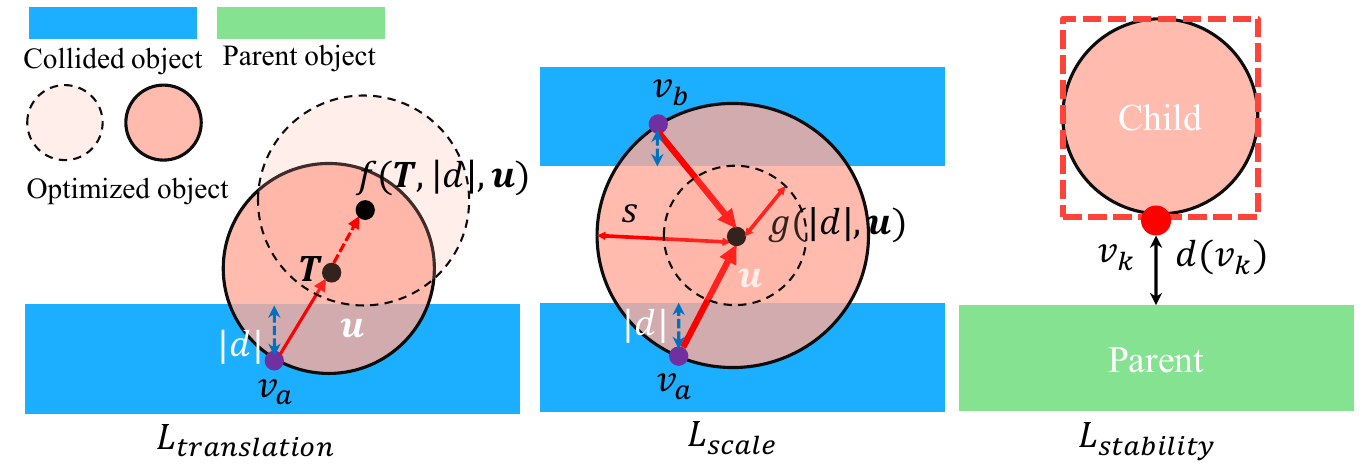}
    \caption{\textit{Collision avoidance} and \textit{stability maintenance}.}\vspace{-6mm}
    \label{fig:physical-illustration} 
\end{figure}
\subsubsection{Physical Plausibility}
Real-world 3D scenes obey physical constraints, ensuring objects remain stable on contact surfaces and collision-free. However, pose alignment with image guidance alone does not guarantee physical plausibility—objects may intersect, float, or sink due to shape discrepancies and errors in scene understanding. See \autoref{fig:ablation} (b) as an example.

Existing methods approximate object geometry using 3D bounding boxes (3DBB)~\cite{yang2024holodeck,yang2024physcene}, which oversimplifies shapes and leads to simplified inter-object relationships.
For example, objects cannot be put within the shelf due to 3D bounding box collision.
This results in simplified scene diversity, especially in tight spaces with complex inter-object relationships (see \autoref{fig:spatial_complaxity_exmaple} for an example).
To address these challenges, we replace 3DBB-based approximations with Signed Distance Fields (SDFs), enabling precise object geometry representation for accurate collision detection and stability constraints. 

The \textbf{physical-aware optimization process} iteratively constructs a SDF-based physical structure, following the scene graph hierarchy: processing the anchor object first to establish a stable foundation, followed by parent and child objects.  
The physics-aware optimization incorporates \textbf{collision} and \textbf{stability constraints}.  
Since retrieved 3D assets are upright, their rotation is constrained to azimuthal adjustments.  

Formally, given a scene graph with $N$ objects, each object has a 5-DoF configuration defined by scale $\scale$, upright rotation $\rotation$, and translation $\translation = (t_x, t_y, t_z)$.  
For computational efficiency, we uniformly sample $n$ points from its triangle surface mesh as its geometric representation and compute its centroid for collision avoidance.

\noindent\textbf{Collision Constraints.}  
We query the scene SDFs using object surface points to detect collision states and define position collision loss $\mathcal{L_{\text{translation}}}$ and scale collision loss $\mathcal{L_{\text{scale}}}$. As shown in~\autoref{fig:physical-illustration}, the deviation caused by collisions impacts translation $T$ as:
\begin{equation}
    \mathcal{L_{\text{translation}}} = \sum_{\mathbf{v}_i \in \mathbf{V}^-}||f(\translation, |d_i|, \mathbf{u}_i) - \translation||_2^2 ,
\end{equation}
where $f(\translation, |d_i|, \mathbf{u}_i) = \translation + \mathbf{u}_i \cdot |d_i|$ computes a collision-free position by adjusting the translation along direction $\mathbf{u}_i$ with step size $|d_i|$.  
Here, $d_i$ is the negative SDF value at a collided point $\mathbf{v}_i$, which belong to the points set with negative SDF $\mathbf{V}^-$ sampled uniformly from the surface.
The direction $\mathbf{u}$ is defined from the collision point toward the model’s centroid, guiding objects away from collisions.  

Collisions also affect object scale $\scale$ due to opposing forces:  
\begin{equation}
    \mathcal{L_{\text{scale}}} = \begin{cases}
    \sum_{\mathbf{v}_i \in \mathbf{V}^-}\bigg(g(|d_i|, \mathbf{u}_i) - \scale\bigg)^2 & \text{if } N_{\text{cluster}} > 1 \\
    0 & \text{otherwise}
    \end{cases},
\end{equation}
where $g(|d_i|, \mathbf{u}_i) = \frac{||\mathbf{u}_i||-|d_i|}{||\mathbf{u}_i||}$ defines the target scale to reduce collision regions.  
$N_{\text{cluster}}$ denotes the number of distinct clusters formed without SDF sign flipping.  
As shown in ~\autoref{fig:physical-illustration}, two surface points $i$ and $j$ with $d_i \leq 0$ and $d_j \leq 0$ belong to different clusters, and thus push the object to be smaller.  

\noindent\textbf{Stability Constraints.}  
Objects are dragged by gravity and rest on their bottom contacting surface.
We ensure stability by enforcing contact between an object’s bottom points and its parent surface, where their SDF values should be zero, as shown in~\autoref{fig:physical-illustration}.  
The stability loss is defined as:

\begin{equation}
    \mathcal{L_{\text{stability}}} = \sum_{\mathbf{v}_i \in \mathbf{V}^B} \bigg(1 -\text{exp}(-d_i^2) \bigg),
\end{equation}
where $\mathbf{V}^{B}$ are the sampled points on the bottom surface of bounding box, and $d_i$ are their corresponding SDF values.  
Further details on collision loss optimization are provided in the \textit{Appendix}.

\subsection{Spatial Coherence Judgment}
After iteratively optimizing object placement, a scene judge powered by GPT-4o evaluates the spatial alignment between the generated 3D scene and the image guidance produced during the layout refinement stage, ensuring consistency in inter-object relationships.

To assess this alignment, we design three metrics:
(1) object category accuracy, comparing the generated scene with the image guidance;
(2) object orientation alignment, measuring how well object orientations match the reference layout;
(3) overall spatial coherence, capturing the holistic consistency of the scene layout.

Each metric is normalized between 0 (lowest) and 1 (highest). If any metric falls below a predefined threshold, the scene judge triggers a re-planning step. Further details are provided in the \textit{Appendix}.

\section{Experiment}

\begin{table*}[t]
  \centering
  \caption{Quantitative evaluation on text–image alignment and visual‑quality preference
           (\textuparrow\,higher is better). \textbf{Bold} marks the best for text control measurement. Visual quality preference indicates GPT-4o and human preference for our method over the baseline. }
  \vspace{-2mm}

  {\footnotesize                
   \setlength{\tabcolsep}{4pt}  

   \begin{tabularx}{\linewidth}{lccc *{4}{C}}
     \toprule
     \multirow{2}{*}{Method}
       & \multicolumn{3}{c}{\textbf{Text–Image Alignment}}
       & \multicolumn{4}{c}{\textbf{Visual‑Quality Preference of Ours (GPT-4o / Human Evaluation)}}\\
     \cmidrule(lr){2-4}\cmidrule(lr){5-8}
       & CLIP↑ & BLIP↑ & VQA↑
       & Object Diversity↑ & Layout Coherence↑ & Spatial Realism↑ & Overall Performance↑ \\
     \midrule
     PhyScene
       & -- & -- & --
       & 80\% / 75\% & 60\% / 46\% & 85\% / 74\% & 50\% / 53\% \\
     DiffuScene
       & 23.11 & 48.28 & 0.7832
       & 75\% / 80\% & 80\% / 90\% & 90\% / 76\% & 80\% / 80\% \\
     SceneTeller
       & 25.27 & 51.99 & 0.7999
       & 80\% / 85\% & 80\% / 71\% & 85\% / 80\% & 80\% / 74\% \\
     Holodeck
       & 28.32 & 46.25 & 0.6815
       & 85\% / 80\% & 83\% / 78\% & 81\% / 86\% & 85\% / 85\% \\
     \textbf{Ours}
       & \textbf{30.71} & \textbf{77.17} & \textbf{0.8269}
       & -- / -- & -- / -- & -- / -- & -- / -- \\
     \bottomrule
   \end{tabularx}
  }
  \label{tab:text_vis}
\end{table*}

\begin{table}[t]
    \centering
    \caption{Physical‑plausibility and interactivity results
             (\textdownarrow\,lower is better for collision/instability).
             \textbf{Bold} indicates the best value.}
    \vspace{-2mm}
    \resizebox{\linewidth}{!}{%
    \begin{tabular}{lcccccc}
        \toprule
        \multirow{2}{*}{Method}
            & \multicolumn{4}{c}{\textbf{Physical Plausibility}}
            & \multicolumn{2}{c}{\textbf{Interactivity}}\\
        \cmidrule(lr){2-5} \cmidrule(lr){6-7}
            & Col‑O↓ & Col‑S↓ & Inst‑O↓ & Inst‑S↓
            & Reach↑ & Walk↑ \\
        \midrule
        PhyScene
            & 17.6\% & 51\% & 18.73\% & 75.22\% & 0.77 & 0.84 \\
        DiffuScene
            & 19.5\% & 55\% & 20.75\% & 83.33\% & 0.74 & 0.83 \\
        SceneTeller
            & 35.2\% & 75\% & 41.17\% & 78.57\% & 0.75 & 0.80 \\
        Holodeck
            & 6.1\% & 21\% & 7.00\% & 31.58\% & 0.90 & \textbf{0.96} \\
        \textbf{Ours}
            & \textbf{0.8\%} & \textbf{6\%} & \textbf{3.20\%} & \textbf{16.67\%}
            & \textbf{0.94} & \textbf{0.96} \\
        \bottomrule
    \end{tabular}}
    \label{tab:phys_interact}
    \vspace{-4mm}
\end{table}

\begin{figure}[t]
    \centering
    \includegraphics[width=\linewidth]{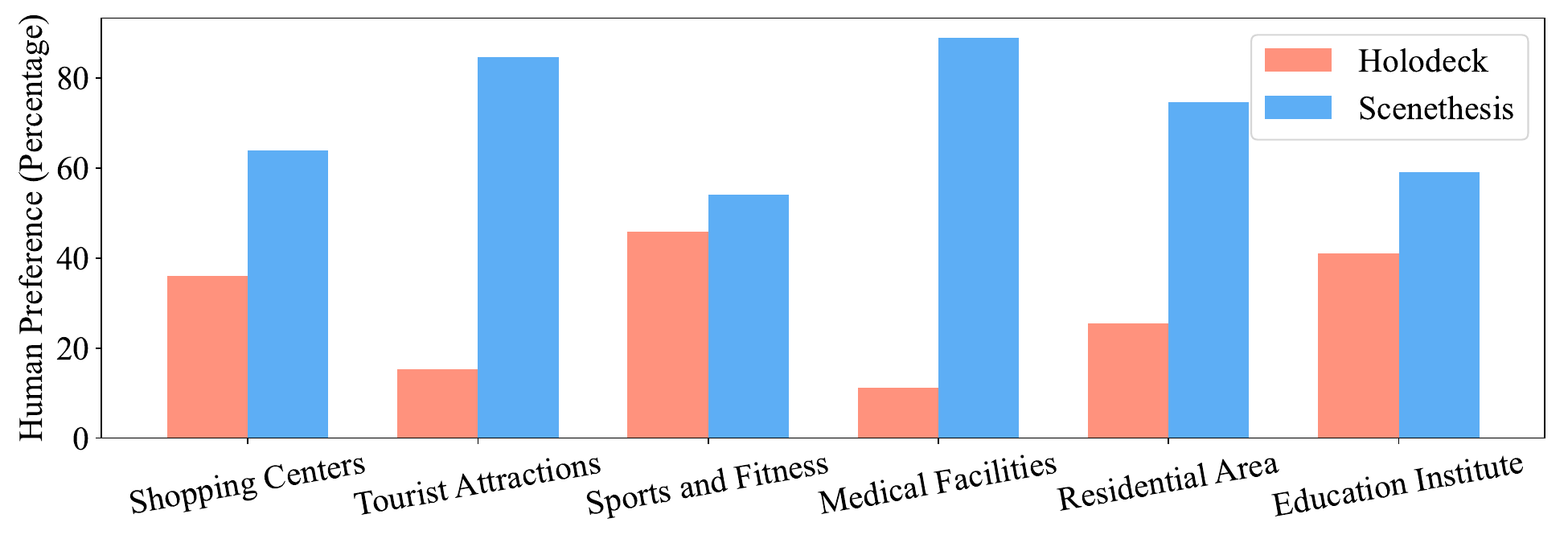}
    \caption{
    Human preference on diverse indoor scenes.} \vspace{-5mm}
    \label{fig:holodeck-compare}
\end{figure}

\begin{figure*}[t]
    \centering
    \includegraphics[width=\linewidth]{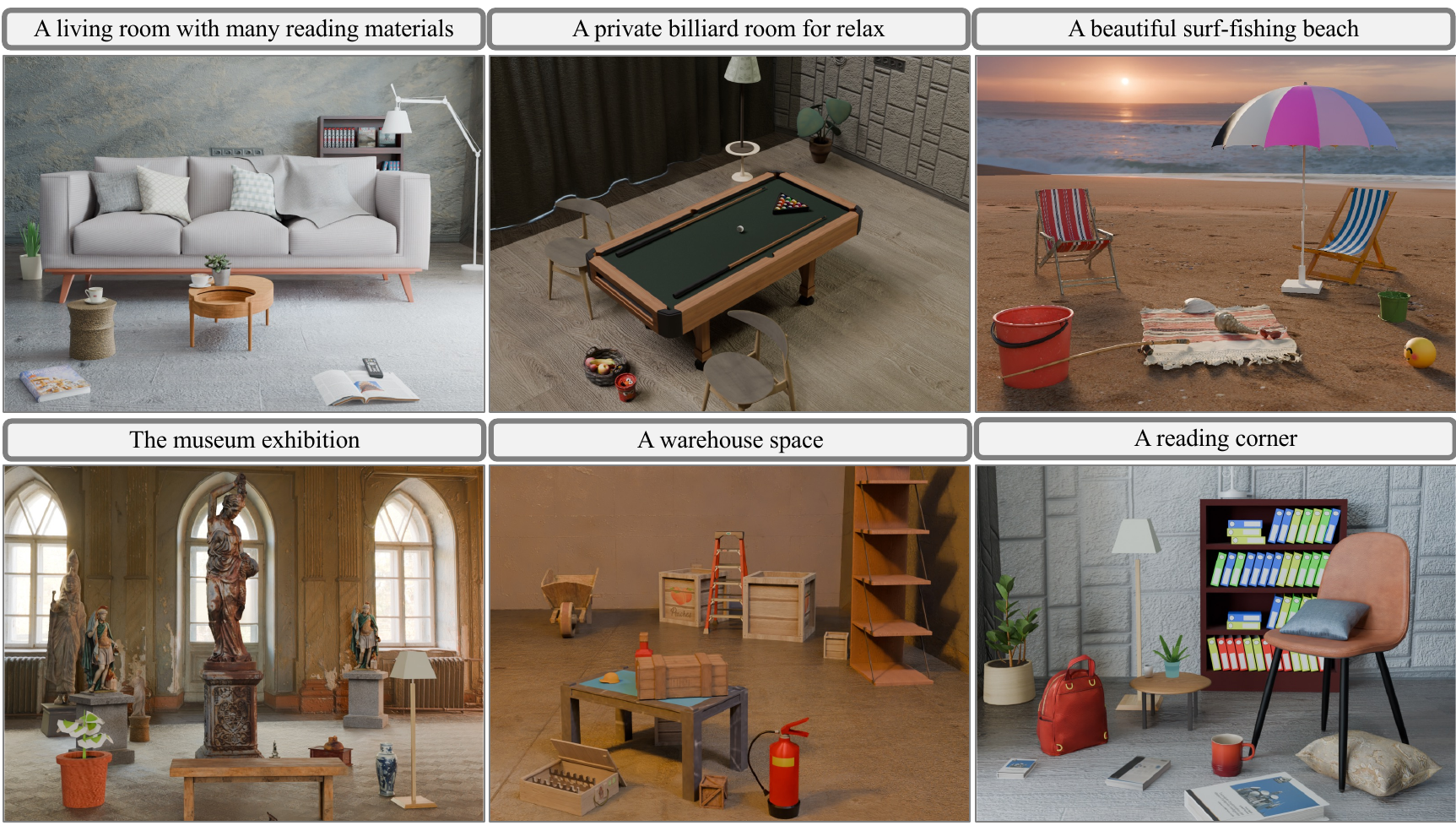}
    \captionof{figure}{ \textbf{Qualitative results of generated indoor and outdoor scenes by \name.} \name\ can generate diverse scenes given user prompts. Visualizations of the scenes at different camera viewpoints can be found in \textit{appendix}.}
    \label{fig:qualitative} \vspace{-4mm}
\end{figure*}

\begin{figure*}[t]
    \centering
    \includegraphics[width=\linewidth]{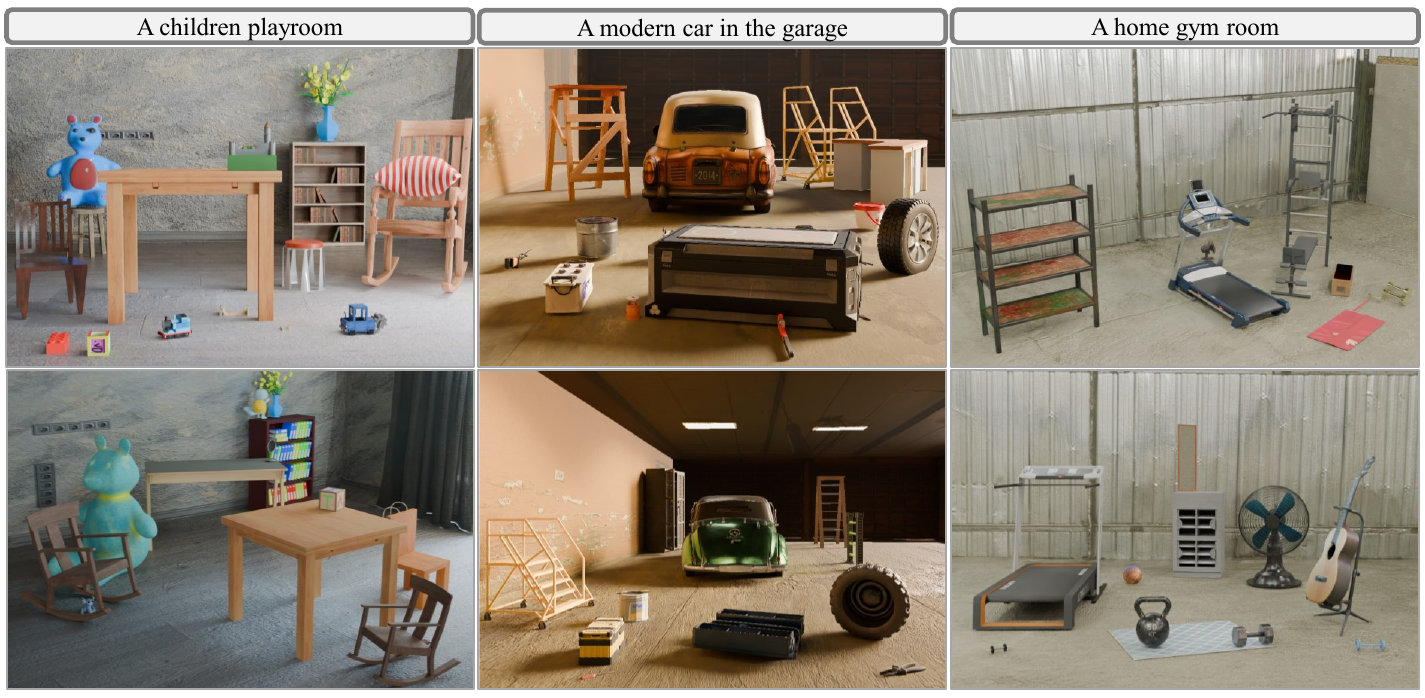}\vspace{-2mm}
    \caption{\textbf{Output Diversity.} Given the same text prompt, \name\ can generate diverse scene with various objects and different layouts. }
    \label{fig:figure3} \vspace{-4mm}
\end{figure*}

\begin{figure}[t]
    \centering
     \includegraphics[width=\linewidth]{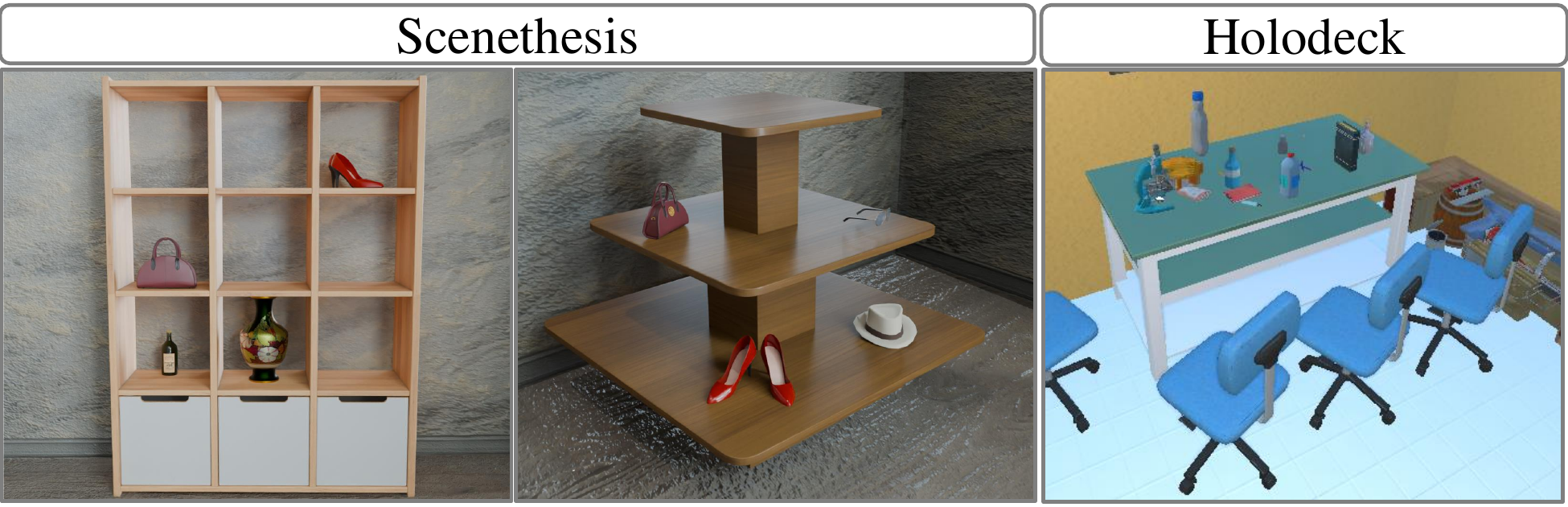}
    \caption{\textbf{Complex spatial realism.} Spatial realism comparison between \name\ and Holodeck. \name\ generates spatially plausible 3D scenes, precisely placing small objects (e.g., bag, wine bottle, shoes, vase) within shelf compartments rather than just on top. This precision, challenging for LLM-based methods, is essential for embodied agent manipulation tasks~\cite{yang2024physcene, nasiriany2024robocasa}.}
    \label{fig:spatial_complaxity_exmaple} \vspace{-3mm}
\end{figure}

\noindent\textbf{Implementation.} 
We use GPT-4o~\cite{hurst2024gpt} as the LLM and image generation in vision module.  
Following Holodeck~\cite{yang2024holodeck}, we retrieve 3D models from a high-quality Objaverse~\cite{deitke2023objaverse} subset. 
Other module details are discussed in the above section. 
The physics-aware optimization is implemented by PyTorch~\cite{paszke2019pytorch} and PyTorch3D~\cite{ravi2020accelerating}. 
Experiments are run on an A100 GPU. 

\noindent\textbf{Baselines.}  
Since we focus on interactive scene generation, methods producing only single-geometry representations are not relevant.  
We compare our approach against open-sourced state-of-the-art (SOTA) \textit{generative methods} (DiffuScene~\cite{tang2024diffuscene}, PhyScene~\cite{yang2024physcene}) and \textit{LLM-based methods} (SceneTeller~\cite{ocal2024sceneteller}, Holodeck~\cite{yang2024holodeck}).  
For fairness, all LLM-based methods use the same ChatGPT version.

\noindent\textbf{Setup.}  
\name\ generates both indoor and outdoor scenes (\autoref{fig:teaser}, \autoref{fig:qualitative}), but for fair comparison, we evaluate only indoor scenes.  
To assess diversity and realism, we generate 22 indoor scenes covering 6 primary and 12 secondary categories from DL3DV-10K~\cite{ling2024dl3dv}:  
\textbf{Residential} (\textit{living room, playroom, garage, warehouse}),  
\textbf{Shopping} (\textit{bookstore, store}),  
\textbf{Tourism} (\textit{museum, piano showroom}),  
\textbf{Sports} (\textit{gym, billiard club}),  
\textbf{Medical} (\textit{ward}),  
\textbf{Education} (\textit{laboratory}).  
Since DiffuScene, PhyScene, and SceneTeller were trained on indoor datasets~\cite{fu20213d} (mainly residential areas), we compare them within this domain.  
Holodeck, which also retrieves models from Objaverse, supports indoor scene generation, enabling comparisons across all indoor categories.  
To mitigate view-dependent bias, we render each scene from two perspectives, yielding 44 image pairs.  
For baselines lacking background generation (e.g. SceneTeller), we render \name\ outputs without an environment map for a fair comparison.

\subsection{Metrics}

We evaluate \textit{controllability} in text-based scene generation methods and three key properties essential for virtual content generation: \textit{layout realism}, \textit{physical plausibility}, and \textit{interactivity}.  

\noindent\textbf{Controllability.}  
Ensuring 3D scene generation aligns with input prompts is crucial. We assess this using:  
(1) \textit{CLIP Score}~\cite{radford2021learning} – cosine similarity between image and text features from CLIP.  
(2) \textit{BLIP Score}~\cite{li2023blip} – image-text alignment using the ITM head of BLIPv2.  
(3) \textit{VQA Score}~\cite{lin2024evaluating} – image-caption alignment based on VQA models.  

\noindent\textbf{Layout Realism.}  
Visual quality and spatial realism are important to reflect real-world scene layouts. We evaluate it using following metrics:  
(1) \textit{Object Diversity} – number of objects and categories in the scene.  
(2) \textit{Layout Coherence} – adherence of object positions and orientations to common sense.  
(3) \textit{Spatial Realism} – presence of diverse spatial relations (e.g., on top of, inside, under).  
(4) \textit{Overall Performance} – alignment of object categories and styles with the scene type.  
Evaluation details and examples are in the \textit{appendix}.

\noindent\textbf{Physical Plausibility.}  
Ensuring object collision-free and stable placement is fundamental for physical simulation environments. We construct the following metrics:  
(1) \textit{Col-O} – average object collision rate,  
(2) \textit{Col-S} – average scene collision rate,  
(3) \textit{Inst-O} – average object instability rate, and  
(4) \textit{Inst-S} – average scene instability rate.  

Collision is tested via mesh-mesh intersections, while instability follows Atlas3D~\cite{chen2024atlas3d}, measured by tracking transformations after physics-based simulation~\cite{warp2022}. These metrics assess scene viability for virtual content creation.  

\noindent\textbf{Interactivity.}  
To ensure objects are accessible and manipulable in the scene based on their functional roles, we follow evaluation metrics from PhyScene~\cite{yang2024physcene}:  
(1) \textit{Reach} – average object reachability rate, and  
(2) \textit{Walk} – ratio of the largest connected walkable area over all walkable regions.

\subsection{Quantitative Evaluation} \label{sec.quantitative}

\noindent\textbf{Controllability.}  
\autoref{tab:text_vis} presents a comprehensive evaluation of text-image alignment. 
Among all baselines, \name\ achieves the highest CLIP, BLIP, and VQA scores, confirming its effectiveness in adhering to text description and the reliability of our agentic pipeline.

\noindent\textbf{Layout Realism.}  
\autoref{tab:text_vis} reports visual realism scores from human evaluations and GPT-4o, a human-aligned evaluator in text-to-3D tasks~\cite{hurst2024gpt,lin2025evaluating}. \name\ achieves SOTA performance on most metrics.  
Despite DiffuScene and PhyScene being trained on dedicated indoor residential datasets~\cite{fu20213d}, the training-free \name\ achieves comparable or superior layout realism in residential areas.  
In broader indoor settings (e.g., shopping centers, tourist attractions, sports facilities), \autoref{tab:text_vis} and \autoref{fig:holodeck-compare} show that \name\ significantly outperforms Holodeck in visual quality and spatial realism.  
These results demonstrate the advantages of visual prior in guiding spatially realistic scene generation.  

\noindent\textbf{Physical Plausibility and Interactivity.}  
\autoref{tab:phys_interact} presents object-level and scene-level physical plausibility metrics, demonstrating that \name\ significantly reduces collisions and enhances stability.  

DiffuScene~\cite{tang2024diffuscene} and SceneTeller~\cite{ocal2024sceneteller}, trained on high-collision datasets~\cite{fu20213d, yang2024physcene}, lack collision detection and stability constraints, leading to frequent object intersections.  
PhyScene~\cite{yang2024physcene} applies physical constraints but inherits dataset-induced collisions.  
Holodeck~\cite{yang2024holodeck} prevents large-object collisions via Depth-First-Search solver but places small objects on predefined surfaces without collision checks, often causing inter-object penetrations (see \textit{appendix}).  
Moreover, none of these baselines address stability, resulting in frequent failures in physics-based simulations.  

In contrast, \name\ integrates physics-aware layout adjustment, ensuring low-collision, stable environments.  
Beyond physical plausibility, \name\ excels in interactivity, achieving superior reachability and walkability scores. These results highlight \name’s ability to generate accessible, navigable environments where objects align with their functional roles and afford interactions.

\subsection{Qualitative Evaluation}
~\autoref{fig:qualitative} showcases diverse scenes generated by \name, demonstrating high fidelity and versatility in both indoor and outdoor environments.  
Compared to LLM-based approaches, \name\ excels in realism and physical plausibility by leveraging image guidance and physics-aware optimization, effectively capturing real-world spatial complexity and diversity.  
\autoref{fig:figure3} presents various 3D layouts generated from the same text prompt, highlighting diverse asset selection and spatial arrangements.  
\name\ supports both simple and detailed prompts—simple prompts enable flexible, user-friendly generation, while detailed prompts allow controllable 3D scene generation (see \textit{appendix}).  

Holodeck restricts small object placement to predefined areas on the top of larger objects.  
In contrast, \name\ enables fine-grained positioning, placing small object at different levels of the support structure (e.g., shelves, carts), as shown in~\autoref{fig:spatial_complaxity_exmaple}.  
LLM-based methods, which lack visual perception, struggle with this level of spatial realism.  
This capability is critical for embodied AI, enabling realistic interactions and meaningful object manipulation in simulation. 
More examples and qualitative comparisons can be found in \textit{appendix}.

{
\setlength\tabcolsep{0.15em} 
\renewcommand{\arraystretch}{0.75} 
\begin{table}[t]
\centering
\caption{Ablation study on the effectiveness of physically plausible optimization. \name\ is the result in ``+Stability'' which includes all constraint components.} \vspace{-2mm}
\ra{1.0}
\resizebox{\linewidth}{!}{%
    \scriptsize %
    \begin{tabular}{l|c|c|c}
    \toprule
    \textit{Component}        & \textit{Pose Alignment} $\uparrow$ & \textit{Collision Rate} $\downarrow$ & \textit{Instability Rate} $\downarrow$ \\ \hline
    Raw layout                & 0.536                                 & 22.7\%                               & 87.3\%                               \\ 
    +Pose Alignment           & 0.732                                 & 10.6\%                               & 74.2\%                               \\ 
    +Collision                & 0.755                                 & 3.6\%                                & 69.8\%                               \\ 
    +Stability              & \textbf{ 0.836}                           & \textbf{ 0.8\% }                         & \textbf{ 3.2\%}                          \\ \bottomrule
    \end{tabular}
}
\label{tab:ablation} \vspace{-4mm} 
\end{table}
} 

\begin{figure}[t]
    \centering
    \includegraphics[width=\linewidth]{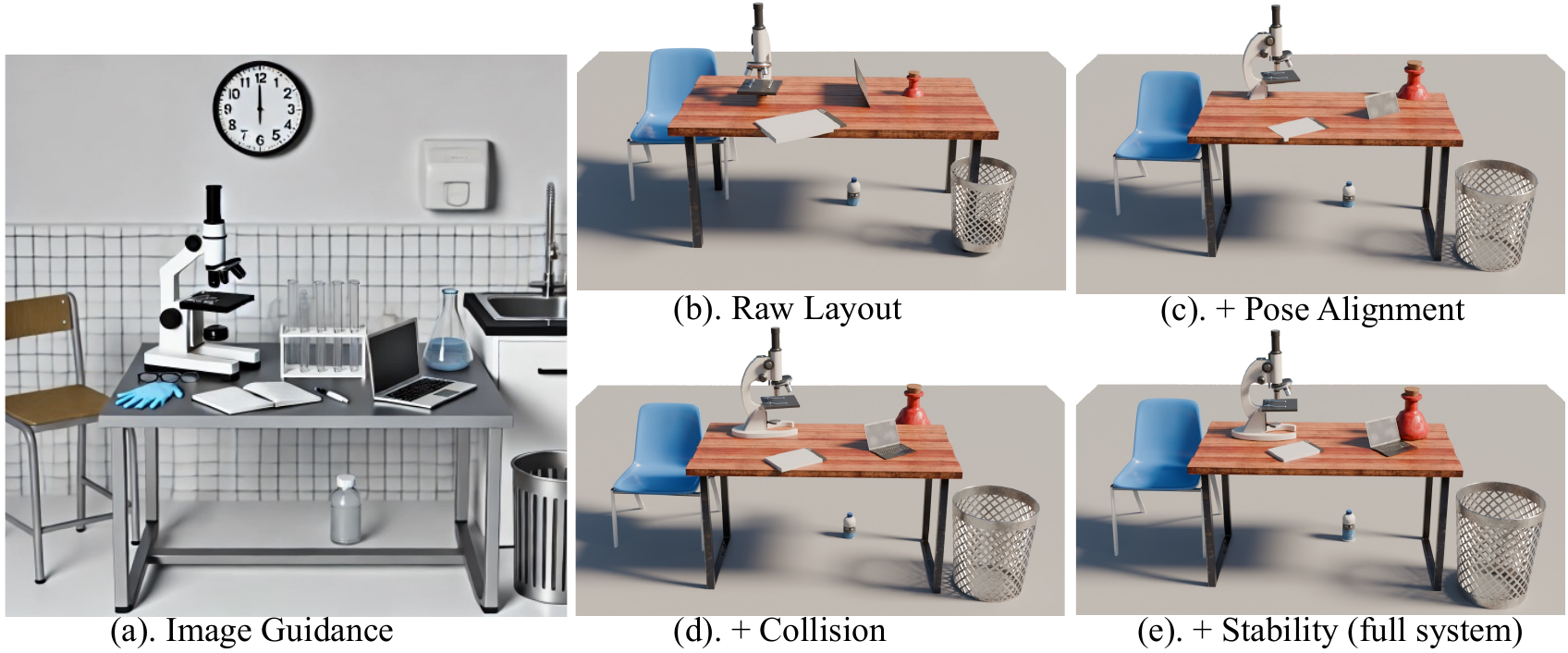}
    \caption{\textbf{Effects of different constraints.}  
(a) \name\ plans the layout and generates image guidance from text input.  
(b) \textit{Raw layout}: places 3D models in estimated 3DBBs.  
(c) \textit{+ Pose alignment}: adjusts 5DoF poses but lacks physical plausibility.  
(d) \textit{+ Collision}: prevents intersections but allows floating objects.  
(e) \textit{+ Stability}: ensures grounded, physically stable objects.}
    \label{fig:ablation} \vspace{-4mm}
\end{figure}
\subsection{Ablation Study}
The physics-aware optimization has three components: \textit{pose alignment}, \textit{collision constraint}, and \textit{stability constraint}. 
We perform ablation studies to assess their effectiveness.  

\noindent\textbf{Metric.}  
For each generated scene, we render the same view as the image guidance and use GPT-4o to assess \textit{pose alignment} based on:  
(1) object orientation, size, and position similarity, and  
(2) spatial coherence of the overall layout.  
The similarity score ranges from 0 to 1, with 1 indicating the highest alignment.  
Wall decorations are ignored in the comparison.  
Additionally, we evaluate object collisions and instability using the method in \autoref{sec.quantitative}.  

\noindent\textbf{Baselines.}  
\textit{Raw Layout}: Objects are placed based on 3DBB estimated by segmentation and depth prediction methods.  
\textit{Pose Alignment}: Aligns object placement with image guidance via correspondence matching.  
\textit{Collision Constraint}: Optimizes placement to avoid collisions.  
\textit{Stability Constraint}: Ensures objects remain stable.  

\noindent\textbf{Results.}  
As shown in \autoref{tab:ablation}, pose alignment significantly improves spatial consistency, while collision and stability constraints enhance physical plausibility, making scenes simulation-ready.  
\autoref{fig:ablation} shows qualitative visualization.

\section{Conclusion and Limitation}
We introduce \name, a training-free agentic framework for generating high-fidelity interactive 3D scenes by leveraging LLM-based coarse scene planning, vision-guided layout refinement, and physics-aware optimization for object position adjustment. A scene judge module ensures spatial coherence. Experimental results demonstrate that it significantly outperforms SOTA baselines in layout coherence, spatial realism, and plausibility. Our approach is limited by retrieval databases since generative 3D methods cannot yet handle articulation. Future advances in generative 3D could overcome this constraint by enabling articulated object synthesis, enhancing scene diversity.

\clearpage
\setcounter{page}{1}
\maketitlesupplementary
\section{Implementation Details of \name}
\subsection{Algorithm Overview}
In this section, we provide a high-level algorithmic overview of \name, with detailed steps outlined in Algorithm~\autoref{alg:text23d}.

\begin{algorithm*}
\caption{Text to 3D Interactive Scene Generation}\label{alg:text23d}
\begin{algorithmic}[1]

\State \textbf{Input:} User text
\State \textbf{Output:} 3D interactive scene layout

\Stage{Stage 1: Coarse Scene Planning }

\State object\_list, upsampled\_prompt $\gets$ \texttt{LLM}(user\_text)  \Comment{obtain the object list and an upsampled prompt}
\EndStage

\vspace{0.5em}
\Stage{Stage 2: Layout Visual Refinement }
\State img\_guidance $\gets$ \texttt{2D\_Diffusion} (upsampled\_prompt) \Comment{generate the guidance image as the reference}
\State cropped\_images $\gets$ \texttt{Grounded\_SAM} (img\_guidance, object\_list)  \Comment{identify each object and crop the images}
\State depth\_map $\gets$ \texttt{Depth\_Pro} (img\_guidance) \Comment{generate depth map}
\State 5DoF\_poses $\gets$ \texttt{Extract\_Poses}(cropped\_images, depth\_map) \Comment{generate initial 5DoF poses}
\State scene\_graph $\gets$ \texttt{VLM} (img\_guidance, object\_list, 5DoF\_poses) \Comment{generate scene graph}

\State 3D\_assets$\gets$ \texttt{CLIP} (cropped\_images, object\_list) \Comment{retrieve 3D assets}
\State environment\_map $\gets$ \texttt{VLM} (upsampled\_prompt) \Comment{retrieve environment maps}
\EndStage

\vspace{0.5em}
\Stage{Stage 3: Physics-aware Optimization}
\State scene\_SDF $\gets$ \texttt{Init\_Scene\_SDF}(anchor\_object) \Comment{compute SDF for each object}

\For{node in scene\_graph.bfs\_traverse()} \Comment{iterate over all objects}
    \State $\scale, \rotation, \translation \gets$ node.pose \Comment{variables to be optimized}
    \State parent\_SDF $\gets$ node.parent.SDF \Comment{obtain parent object's SDF}
    
    \For{iteration = 1 to max\_iterations}
        \State mesh $\gets$ \texttt{Get\_Object\_Mesh}(node)
        \State mesh$^*$ $\gets$ \texttt{Apply\_Transform}(mesh, $\scale$, $\rotation$, $\translation$) \Comment{coordinate alignment}

        \Statex
        \State img\_rendered, depth\_rendered $\gets$ \texttt{Render}(mesh$^*$, camera) \Comment{render RGB and depth images}

        \Statex
        \State correspondence $\gets$ \texttt{RoMa}(img\_guidance, img\_rendered) \Comment{correspondence matching}
        \State mesh\_points $\gets$ \texttt{Get\_Point\_Clouds}(depth\_rendered, correspondence, camera)
        \State guided\_points $\gets$ \texttt{Get\_Point\_Clouds}(depth\_map, correspondence, camera)

        \Statex
        \State $L_{pose\_2D}$ $\gets$ \texttt{Dist\_2D}(correspondence) \Comment{loss computation}
        \State $L_{pose\_3D}$ $\gets$ \texttt{Dist\_3D}(mesh\_points, guided\_points)
        \State $L_{collision}$ $\gets$ \texttt{Collision}(mesh$^*$, scene\_SDF)
        \State $L_{stability}$ $\gets$ \texttt{Stability}(bottom\_points(mesh), parent\_SDF)
        
        \State $loss \gets \lambda \mathcal{L}_{pose}+ \lambda_{collision} \mathcal{L}_{collision} + \lambda_{stability} \mathcal{L}_{stability}$
        \State $loss.\texttt{Backward()}$

        \Statex
        \State $optimizer.$\texttt{Step()} \Comment{pose optimization}
        \State $optimizer.$\texttt{Zero\_Grad()}
    \EndFor
    \State scene\_SDF $\gets$ \texttt{Update\_Scene\_SDF}(scene\_SDF, node)
\EndFor
\EndStage

\vspace{0.5em}
\Stage{Stage 4: Scene Spatial Coherent Judgment}
\State Multi-view images $\gets$ \texttt{Render} (optimized\_3D\_scene)
\State Qualified $\gets$ \texttt{VLM} (Multi-view images)

\If{\textbf{not} qualified}               
    \State \textbf{goto Stage 1}          \Comment{re-generate if current optimization fails.}
\EndIf

\vspace{0.5em}
\EndStage

\State \textbf{Return:} Optimized 3D interactive scene
\end{algorithmic}
\end{algorithm*}

\subsection{Method Details}
\subsubsection{Coarse Scene Planning} 
Using the user's scene prompt as input, the LLM (powered by GPT-4o~\cite{hurst2024gpt}) follows a six-step process:
\begin{enumerate}
    \item Interpreting the user's scene prompt.
    \item Reviewing the object categories available in the provided asset database.
    \item Selecting relevant objects from the asset list.
    \item Cross-checking the availability of the selected objects.
    \item Planning the scene using the selected objects.
    \item Generating output files according to the specified standards.
\end{enumerate}

The final coarse scene planning output consists of two components: a list of selected object categories commonly found in the scene (defining anchor object and other common objects) and an upsampled prompt that outlines the scene's spatial hierarchy. The designed prompt presents in \textit{Coarse Scene Planning Instruction Prompts}~\autoref{box:coarse-instruction} and the output example is in \textit{Coarse Scene Planning Output Example}~\autoref{box:coarse-scene-planning-output-example}.

\subsubsection{Layout Visual Refinement} 
Based on the upsampled prompt, GPT-4o generates an image to serve as fine-grained layout guidance. Several post-processing steps are applied to the generated image:
\begin{itemize}
    \item Scene Graph Construction: GPT-4o~\cite{hurst2024gpt} is used to generate a scene graph, defining the ground as the \textit{root object}, along with \textit{parent objects} and their corresponding \textit{child objects}. Additionally, Grounded-SAM~\cite{ren2024grounded} segments each object in the image to obtain masks and cropped images. These are then projected into 3D space using Depth Pro~\cite{bochkovskii2024depth}, allowing for the initial positioning of objects within a spatial relationship graph.
    \item Asset Retrieval. CLIP (ViT-L/14 trianed on LAION-2B) image and semantic features are employed to retrieve 3D assets that align with the image guidance. GPT-4o~\cite{hurst2024gpt} is further utilized to select the most relevant environment map based on the upsampled prompt. It is important to note that \name\ focuses on layout planning for objects on the ground, while background elements, such as wall decorations, lighting, or outdoor settings (e.g., sunshine or the sea), are visually determined by the environment map.
\end{itemize}
The output of the fine-grained layout planning includes the generated image as guidance, a scene graph with the initial poses of the objects, the retrieved assets, and the retrieved environment map. The visual details are presented in the \textit{video}.

\subsubsection{Physics-aware optimization Details} The physics-aware optimization is an iterative optimization process that consists of two key components: \textbf{pose alignment} optimization and \textbf{physical plausibility} optimization. pose alignment optimization focuses on aligning the position, dimension, and orientation of 3D models with their counterparts in the image guidance to ensure visual coherence for spatial relationships. Physical plausibility optimization ensures that the 3D models in the scene are free from collisions and maintain stability, contributing to a realistic and physically consistent layout.

\begin{figure*}[t]
    \centering
    \begin{tabular}{ccc}
    \includegraphics[width=0.33\linewidth]{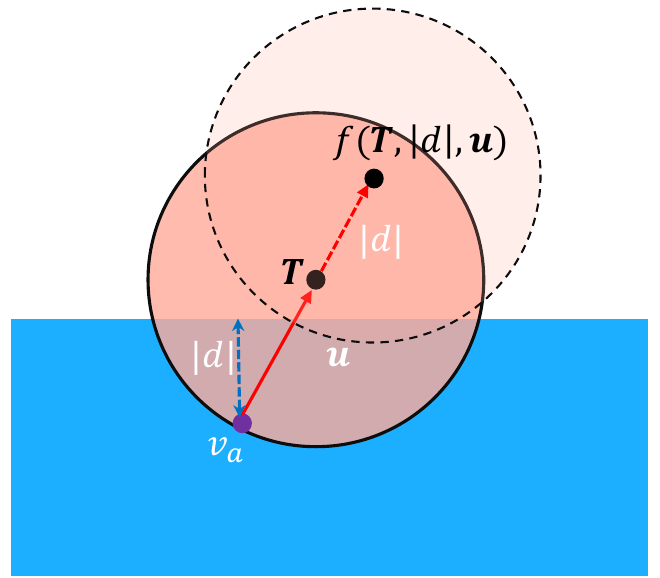} &
    \includegraphics[width=0.33\linewidth]{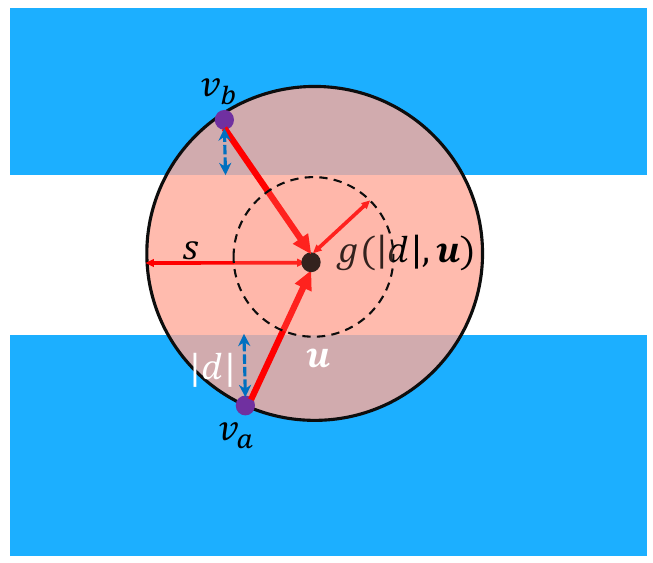} &
    \includegraphics[width=0.33\linewidth]{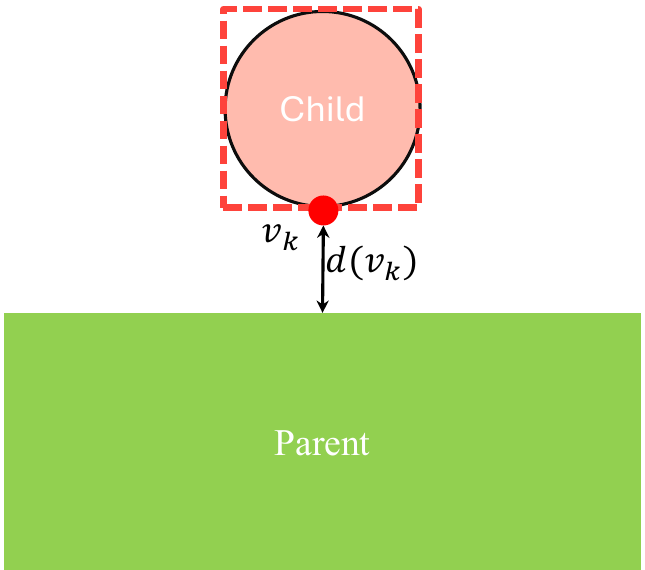} \\
    (a). $\mathcal{L_{\text{translation}}}$ & (b). $\mathcal{L_{\text{scale}}}$ & (c). $\mathcal{L_{\text{stability}}}$
    \end{tabular}
    \caption{Illustration of \textit{collision avoidance} and \textit{stability maintenance}.  The solid-line circle indicates the 3D object's current position, while the dotted-line circle marks its anticipated position. The black dot represents the centroid of the target object, the purple dots indicate surface nodes with negative SDF values, and the red point $v_k$ is the bottom node of the object. (a). The collision pushes the circle object out of the rectangle along the direction from the sampled point to the circle's center by step $|d|$. (b). The collision indicates the object is too large and negative signed distance fields (SDF) points (\ie point $v_a$ and point $v_b$) are detected from distinct classes of the object during optimization. The collision loss shrinks the object size such that there are no different clusters of negative SDF points on the object surface can be detected. (c). The \textit{stability maintenance} keeps the the child and the parent to be as close as possible.}
    \label{fig:illustration-physics}
\end{figure*}

\paragraph{Pose Alignment.} To align the position, dimension, and orientation for objects in rendered image and their counterpart in image guidance, \name\ applies the dense semantic correspondence matching from RoMa~\cite{edstedt2024roma}. That is, minimizing the distance between correspondence points in the rendered image $I$ and the guided image $\Tilde{I}$. Suppose there are $N$ objects in the rendered image $I$, each represented by $\mathbf{o}$ and defined by a 5-DoF configuration, which includes scale $\scale$, upright rotation $\rotation$, and translation $\translation = (t_x, t_y, t_z)$. The counterpart of each object in the generated image $\Tilde{I}$ is denoted as $\mathbf{\Tilde{o}}$. The objective of ensuring visual coherence is to minimize the distance between corresponding points by optimizing the 5-DoF parameters. This ensures that the spatial positions, dimension, and orientations of the 3D models are closely aligned with their counterparts in the guided image. The matching process is formalized as:

\begin{equation}
    \{ p(x,y), \tilde{p}(x,y) \}_i^m = \text{RoMa}(\object, \objectref),
\end{equation}
where $p(x,y), \tilde{p}(x,y)$ are correspondent pair in object $\mathbf{o}$ and $\Tilde{o}$. We select $m$ pair points in each optimization iteration with confident score higher than $\tau$. The higher confidence score indicates a higher probability of matching. We minimize the 2D pixel distance and 3D projected point clouds distance between the matched pair denoted as follows: 
\begin{align}
    \mathcal{L}_{pose} = \lambda_{2d} \mathcal{L}_{2d} + \lambda_{3d} \mathcal{L}_{3d},
\end{align}
where $\lambda_{2d}$ and $\lambda_{3d}$ are coefficients of the 2D pixel loss and 3D point cloud loss denoted as $\mathcal{L}_{2d}$ and $\mathcal{L}_{3d}$. 
\newline

\paragraph{Physical Plausibility.} Physical plausibility ensure generated 3D scenes adhering to fundamental physical principles. Instead of using 3D bounding box (3DBB) as object approximation, \name\ accurately detects collision state from the surface points of the 3D models using signed distance field (SDF). The \textit{collision avoidance} and \textit{stability maintenance} as illustrated in \autoref{fig:illustration-physics}. 

The \textit{collision avoidance} affects the translation $T$ by:
\begin{equation}
    \mathcal{L_{\text{translation}}} = ||f(\translation, |d|, \mathbf{u}) - \translation||_2^2 ,
\end{equation}
where $f(\translation, d, \mathbf{u}) = T + \mathbf{u} \cdot |d|$ computes a collision-free position $\hat{\translation}$ by adjusting $\translation$ along direction $\mathbf{u}$ with step size $d$.  
Here, $d$ is the negative SDF value at a collided point $v_i$ such that $d(v_i) \leq 0$ and $|d| = \max(0, -d(v_i))$ is the negative SDF value $d$ after being processed through a ReLU function, meaning only collided points contribute to this collision term. The direction $\mathbf{u}$ is defined from the collision point toward the model’s centroid $\centroid$, guiding objects away from collisions. 

The \textit{collision avoidance} affects the scaling $\scale$ by detecting that object collides from at least two different directions:
\begin{equation}
    \mathcal{L_{\text{scale}}} = \begin{cases}
    \sum_{\mathbf{v}_i \in \mathbf{V}^-}\bigg(g(|d_i|, \mathbf{u}_i) - \scale\bigg)^2 & \text{if } N_{\text{cluster}} > 1 \\
    0 & \text{otherwise}
    \end{cases},
\end{equation}
where $g(|d_i|, \mathbf{u}_i) = \frac{||\mathbf{u}_i||-|d_i|}{||\mathbf{u}_i||}$ defines the target scale to reduce collision regions.  
$N_{\text{cluster}}$ denotes the number of distinct clusters formed without SDF sign flipping.  
As shown in ~\autoref{fig:physical-illustration}, two surface points $i$ and $j$ with $d_i \leq 0$ and $d_j \leq 0$ belong to different clusters, and thus push the object to be smaller.

The \textit{stability maintenance} affects the translation $T$ by:
\begin{equation}
    \mathcal{L_{\text{stability}}} = \sum_{\mathbf{v}_i \in \mathbf{V}^B} \bigg(1 -\text{exp}(-d_i^2) \bigg),
\end{equation}
where $\mathbf{V}^{B}$ are the sampled points on the bottom surface of bounding box, and $d_i$ are their corresponding SDF values.

\paragraph{Method Overview}
Building on the physics-aware optimization described above, we now integrate pose spatial constraints and physical constraints into the text-to-3D optimization framework. Since physical loss depends on object geometry and can alter its spatial position, it may affect the rendered visible regions due to occlusions or shifts in the scene, introducing biases in pose alignment when using image guidance for semantic correspondence matching. To mitigate this, we adopt a two-stage optimization strategy: first, we optimize pose alignment based on correspondence matching; then, we refine object placement with physical constraints to ensure a visually coherent and physically plausible 3D scene. The following function defines the joint optimization of object position, orientation, and scale:
\begin{equation}
    \mathcal{L} = \lambda_{p} \mathcal{L}_{pose} + \lambda_{c\_T} \mathcal{L_{\text{translation}}} +\lambda_{c\_S} \mathcal{L_{\text{scale}}} + \lambda_{s} \mathcal{L_{\text{stability}}}
\end{equation}

\begin{figure}
    \centering
    \includegraphics[width=\linewidth]{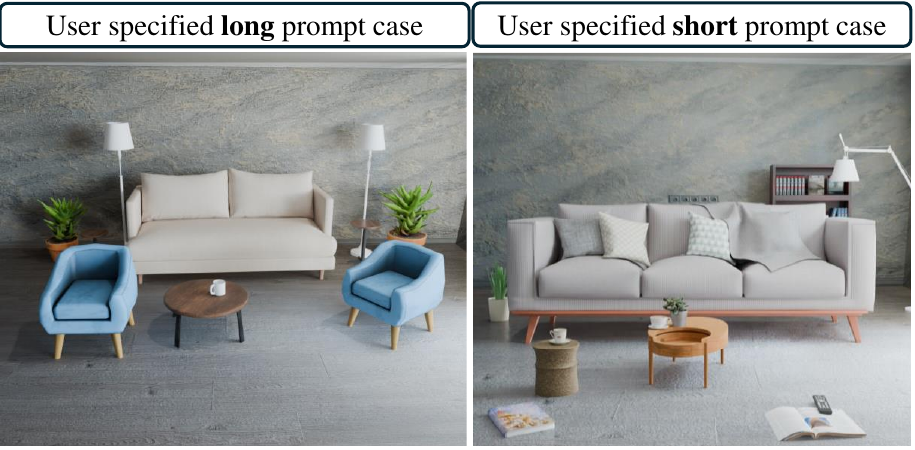}
    \caption{\textit{Short prompt}: a living room with reading materials; \textit{detailed long prompt}: A living room that provide a neutral and cozy space with a minimalist design. At the center of the scene, a light beige sofa is positioned against a textured stone wall in the background. In front of the sofa, a round wooden coffee table sits on the floor, with a white coffee cup placed on top. Two blue armchairs are symmetrically arranged on either side of the coffee table, facing inward toward the sofa. Behind each armchair, a tall white floor lamp stands, providing ambient lighting. Next to the lamps, green potted plants are placed near the wall, adding a natural decorative touch.}
    \label{fig:short_long_promts}
\end{figure}

\subsection{Experiment Details}
\paragraph{Parameters.}
For pose alignment, we select the $m=100$ correspondence points with matching conference $\tau \geq 0.6$ in each optimization iteration. Additionally, we uniformly select $n=400$ samples from the surface of 3D model to accurately detect the collision and stability states in each optimization iteration. 

We explored \adam and \SGD as the optimizer during the optimization process.
Though \adam has been widely applied for training deep neural networks, the adaptive momentum makes the optimization unstable, leading to sub-optimal optimized pose.  
So we use \SGD in our implementation. The optimization implementation is based on pytorch3D~\cite{ravi2020accelerating} and the visualization is rendered using Blender.

\paragraph{Prompts.} \name\ supports both short and detailed user-specified prompts. A short prompt provides a user-friendly and flexible approach to 3D scene generation, where the LLM interprets the input, revisits the available 3D models in database, selects the common objects and anchor objects, and generates an upsampled text prompt for coarse layout planning.
In contrast, a long prompt, which includes user-defined objects and inter-object relationships, enables greater user control over 3D scene generation. In this case, the LLM directly reasons over the detailed prompt, revisits available 3D models in the database, and defines the anchor object, skipping the upsampling stage.
We illustrate examples of short and long prompts defining a living room in ~\autoref{fig:short_long_promts}.

We compared four baselines—Physcene~\cite{yang2024physcene}, Diffuscene~\cite{tang2024diffuscene}, SceneTeller~\cite{ocal2024sceneteller}, and Holodeck~\cite{yang2024holodeck}—evaluating visual quality, physical plausibility, and interactivity metrics. Among them, Diffuscene, SceneTeller, and Holodeck perform text-to-3D scene generation.  
For visual quality assessment, we use both a user study and GPT-4o as evaluation tools. Unlike other baselines, which generate only living room, bedroom, and dining room scenes from 3D-FRONT~\cite{fu20213d}, Holodeck and \name\ utilize Objaverse~\cite{deitke2023objaverse} as a retrieval database, enabling more diverse indoor scene generation.  

We outline the GPT-4o prompt assessment for both baseline evaluation and ablation evaluation as follows:

\begin{figure}[t]
    \centering
    \includegraphics[width=\linewidth]{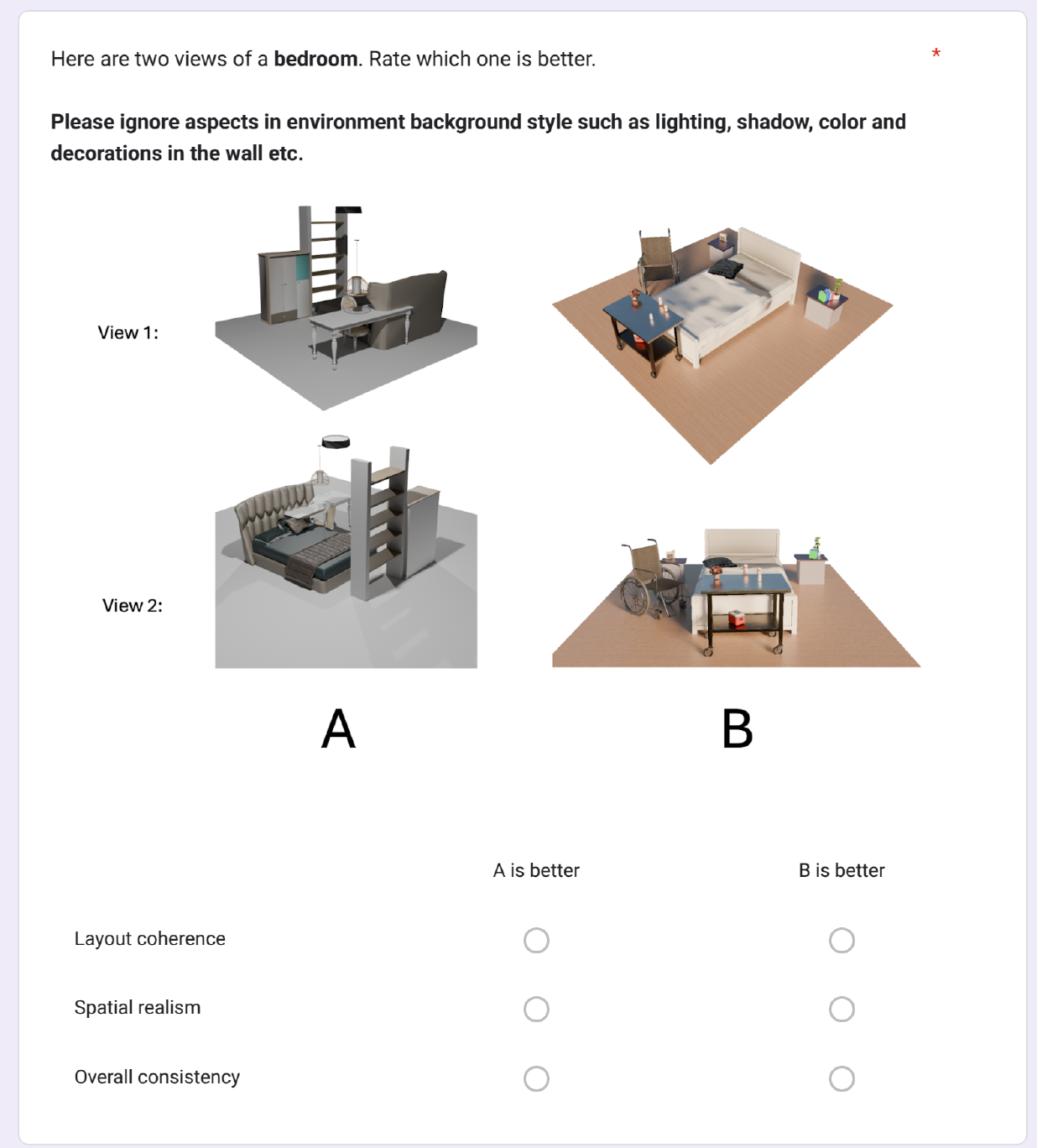}
    \caption{\textbf{User study example.}}
    \label{fig:user_study_example}
\end{figure}

\begin{figure}[t]
    \centering
    \includegraphics[width=\linewidth]{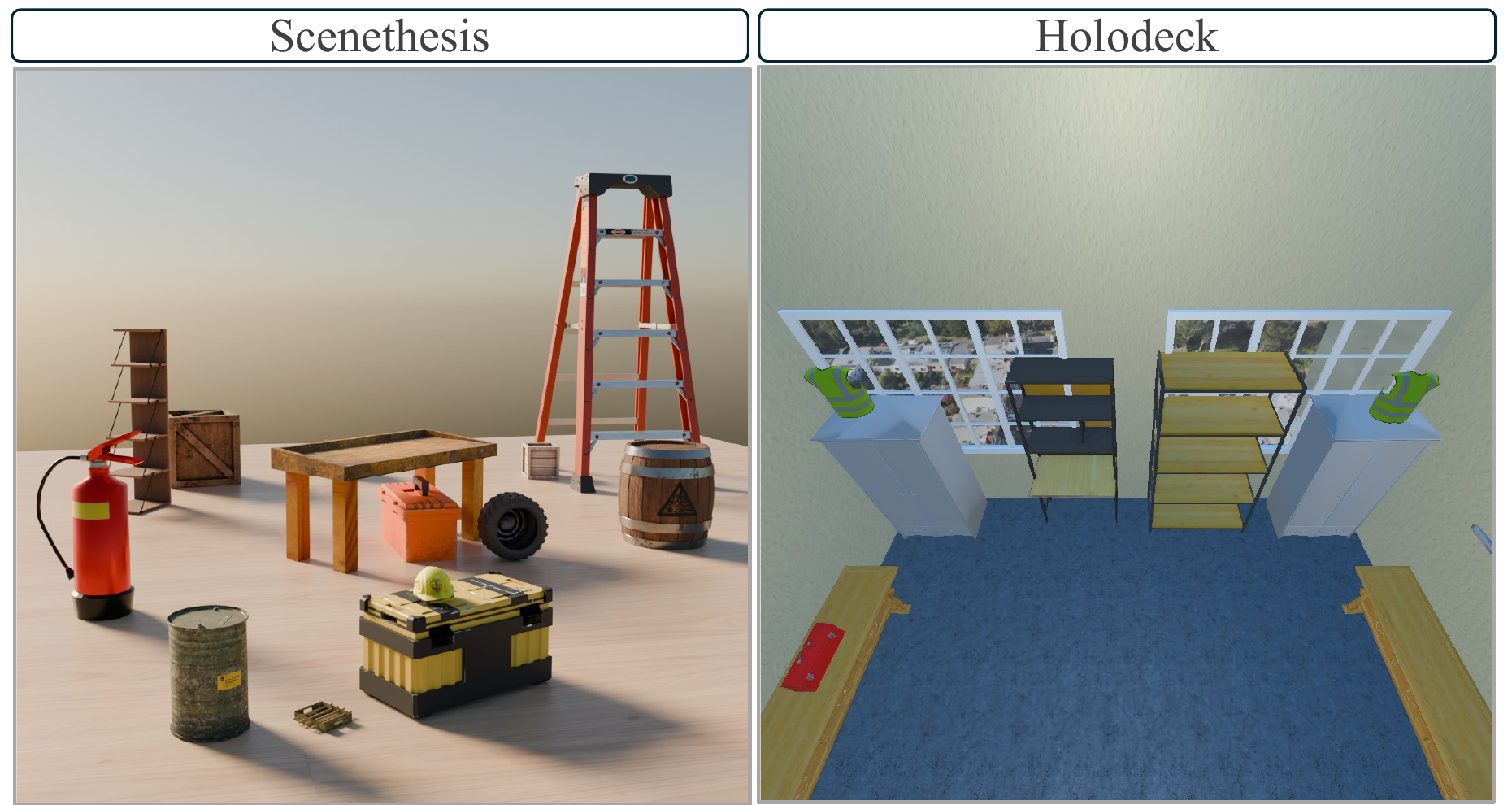}
    \caption{An example comparison of generated scenes given user prompt: ``a warehouse''. Note that \name's scenes are rendered without an environment map to ensure a fair comparison with Holodeck's scenes.}
    \label{fig:scene_comparison}\vspace{-2mm}
\end{figure}

\begin{itemize}
    \item \textbf{Comparison with baselines by GPT-4o:} GPT-4o is employed to evaluate the generated scenes for four metrics: \textit{object diversity}, \textit{layout coherence}, \textit{spatial realism and complexity}, and \textit{overall performance}. The evaluation prompts are detailed in the \textit{Instruction Prompts for Evaluating Generated Scene} ~\autoref{box:scene-evaluation}. Additionally, a comparison example of the generated scenes is provided in \autoref{fig:scene_comparison} with their evaluation results generated by GPT-4o detailed in \textit{Evaluation Example of Generated Scenes} ~\autoref{box:scene_evaluation_results}. 
    \item {\textbf{Comparison with baselines by human preference}}: We applied a user study to study human preference of baseline method with our method. See \autoref{fig:user_study_example} as an example. There are 69 users took our survey. 
    \item \textbf{Evaluation in Ablation Studies:} GPT-4o is also utilized to assess the pose alignment metric during the ablation studies of \name's physics-aware optimization. This evaluation measures the similarity of object position, size, and orientation with their counterparts in the image guidance, as well as the overall visual coherence of the layout. The instructions for assessing pose alignment are provided in the \textit{Instruction Prompts for Ablation Study}~\autoref{box:spatial_alignment}.
\end{itemize}

\begin{figure}[t]
    \centering
    \includegraphics[width=\linewidth]{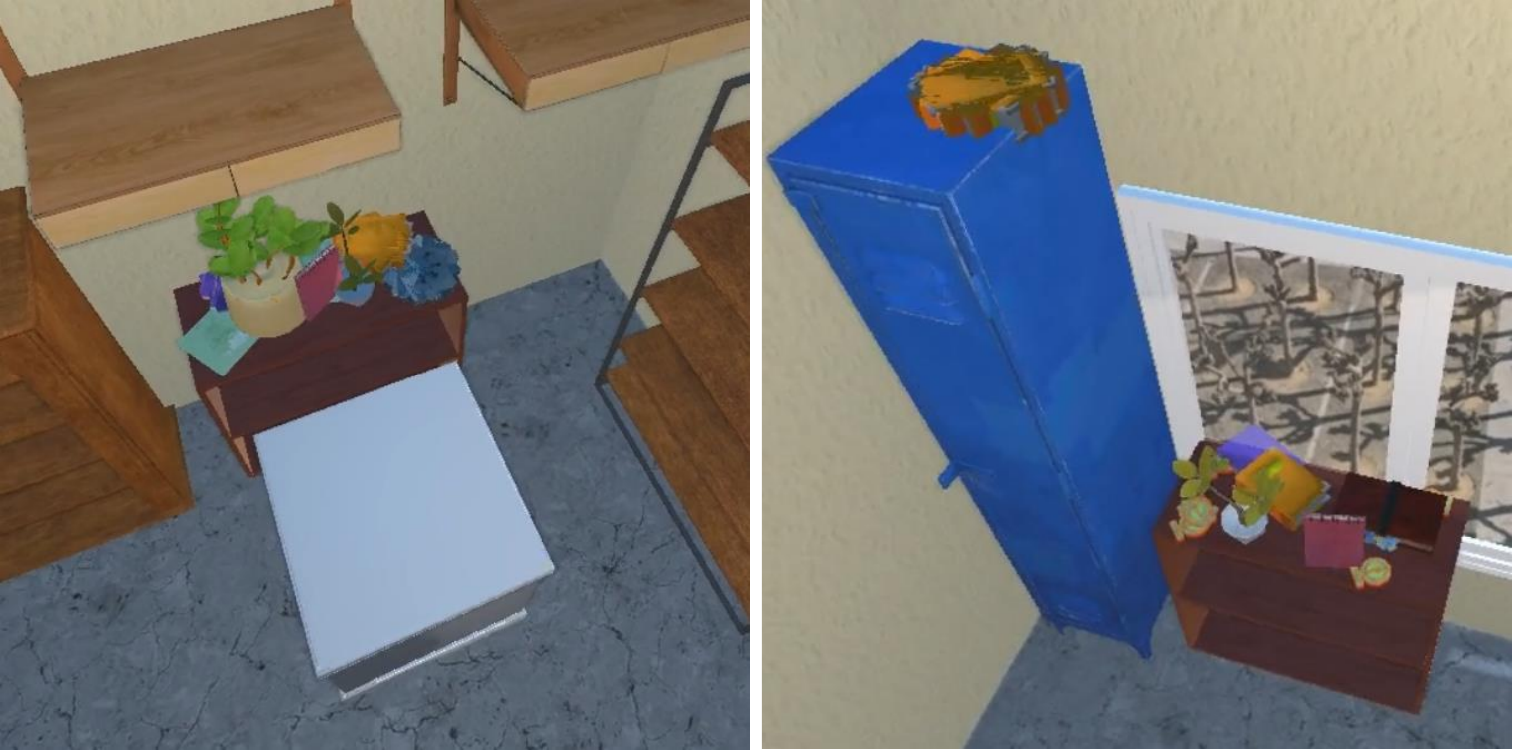}
    \caption{An example of objects collision from Holodeck's scenes.}
    \label{fig:holodeck_collide}\vspace{-2mm}
\end{figure}

\paragraph{Results.} We present additional qualitative results for \name's scenes.

\begin{figure*}[t]
    \centering
    \includegraphics[width=\linewidth]{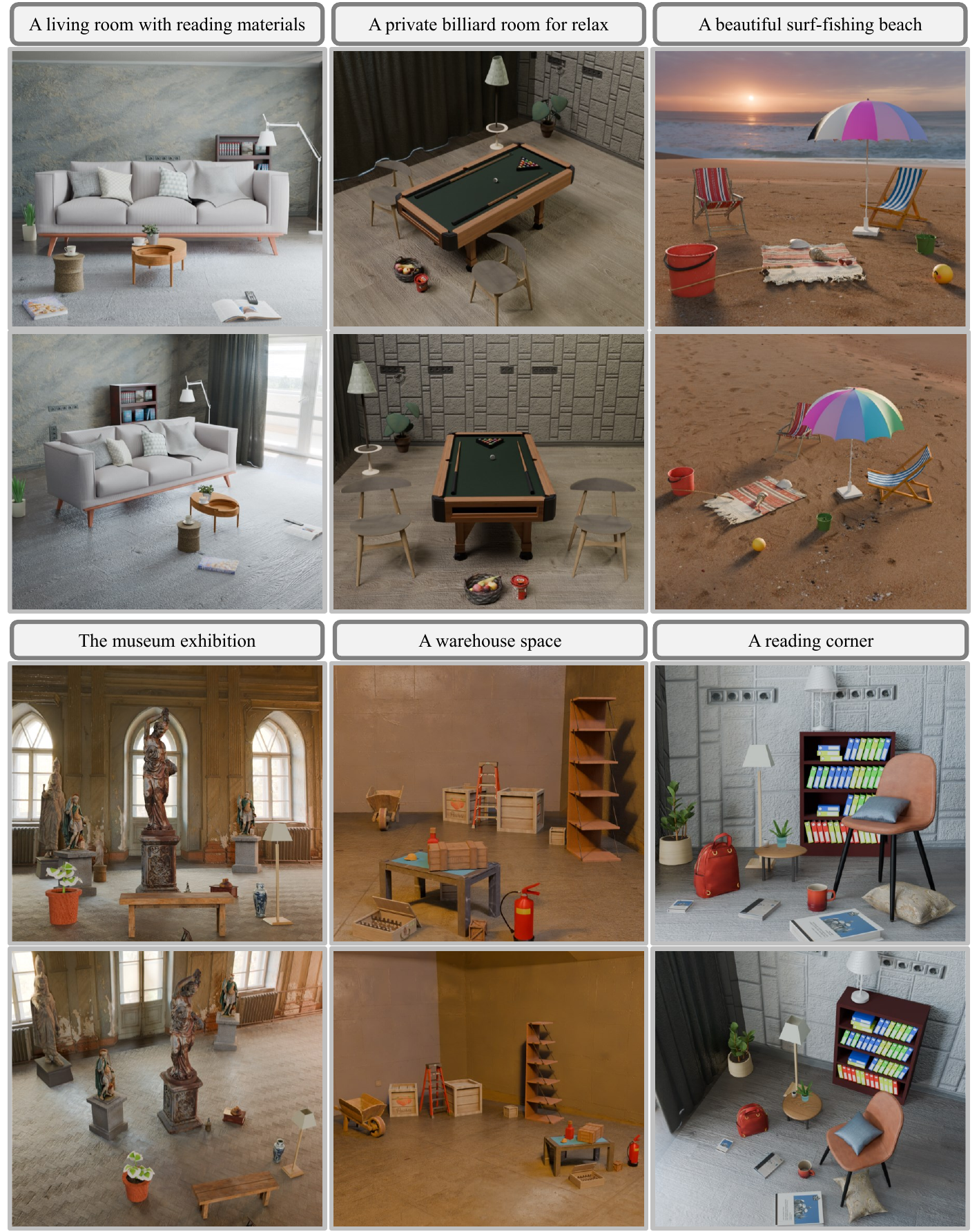}
    \caption{Qualitative results of generated indoor and outdoor scenes by \name\ at different camera viewpoints}
    \label{fig:qualitative_mutliview}
\end{figure*}

\begin{figure*}[t]
    \centering
    \includegraphics[width=\linewidth]{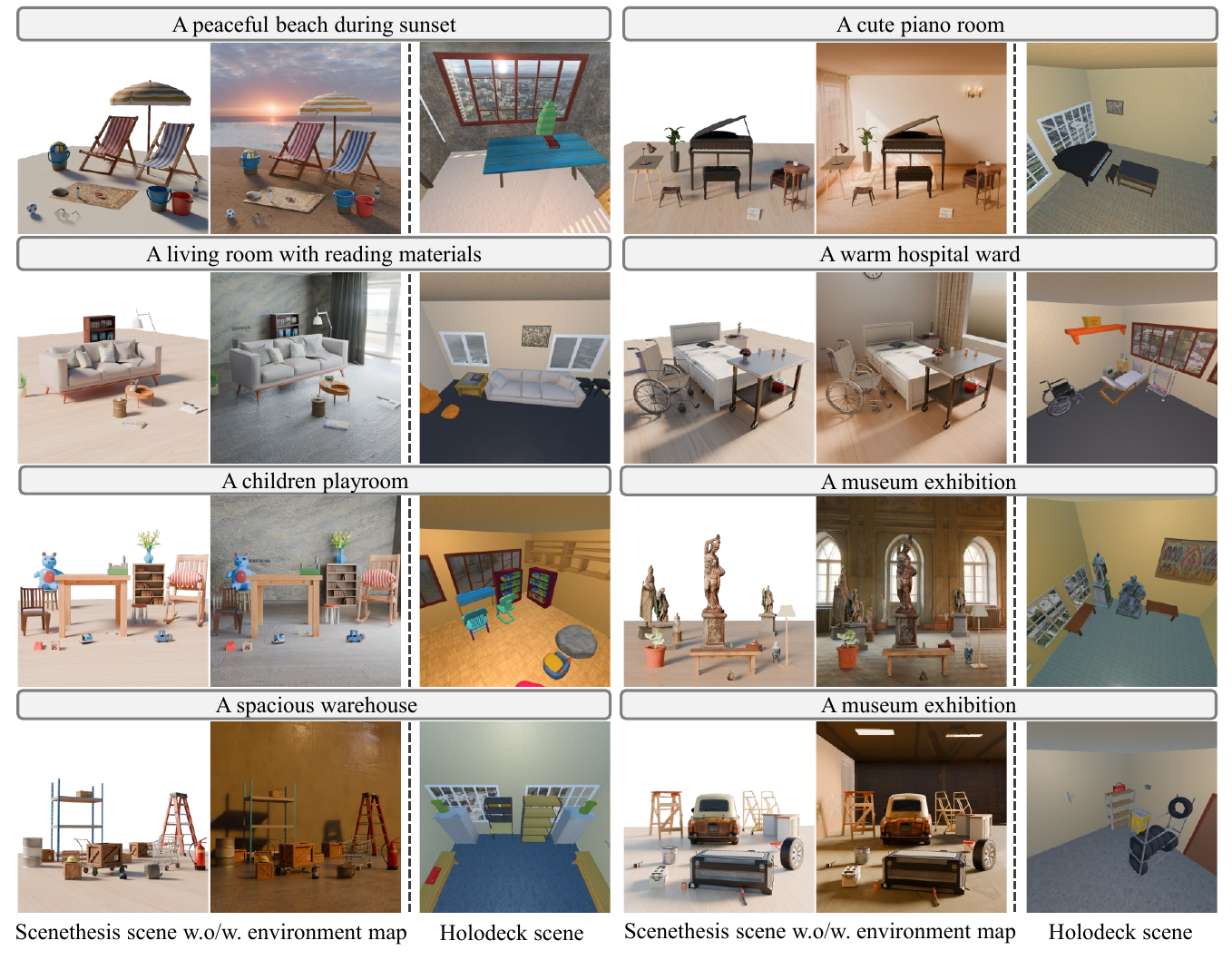}
    \caption{Visualization comparison of generated scenes between \name\ and Holodeck. The first column of images shows scenes generated by \name\ without an environment map, the second column displays scenes generated by \name\ with an environment map, and the third column presents scenes generated by Holodeck. The evaluation metrics, including \textit{object diversity}, \textit{layout coherence}, \textit{spatial realism}, and \textit{overall performance}, are detailed in the \textbf{Experiment} section. \name's scenes have a wider variety of spatial relationships, such as \textit{``on top of''}, \textit{``inside''}, and \textit{``under''}, compared to those generated by Holodeck~\cite{yang2024holodeck}, which supports only \textit{``on top of''} spatial relation.
    In addition, Holodeck lacks visual perception and usually generates misoriented objects, e.g. shelves occlude the window in children playroom and warehouse, chair orients towards the window in the hospital case, hindering their functionalities.  
    }
    \label{fig:holodeck_comp}
\end{figure*}

\begin{figure*}[t]
    \centering
    \includegraphics[width=\linewidth]{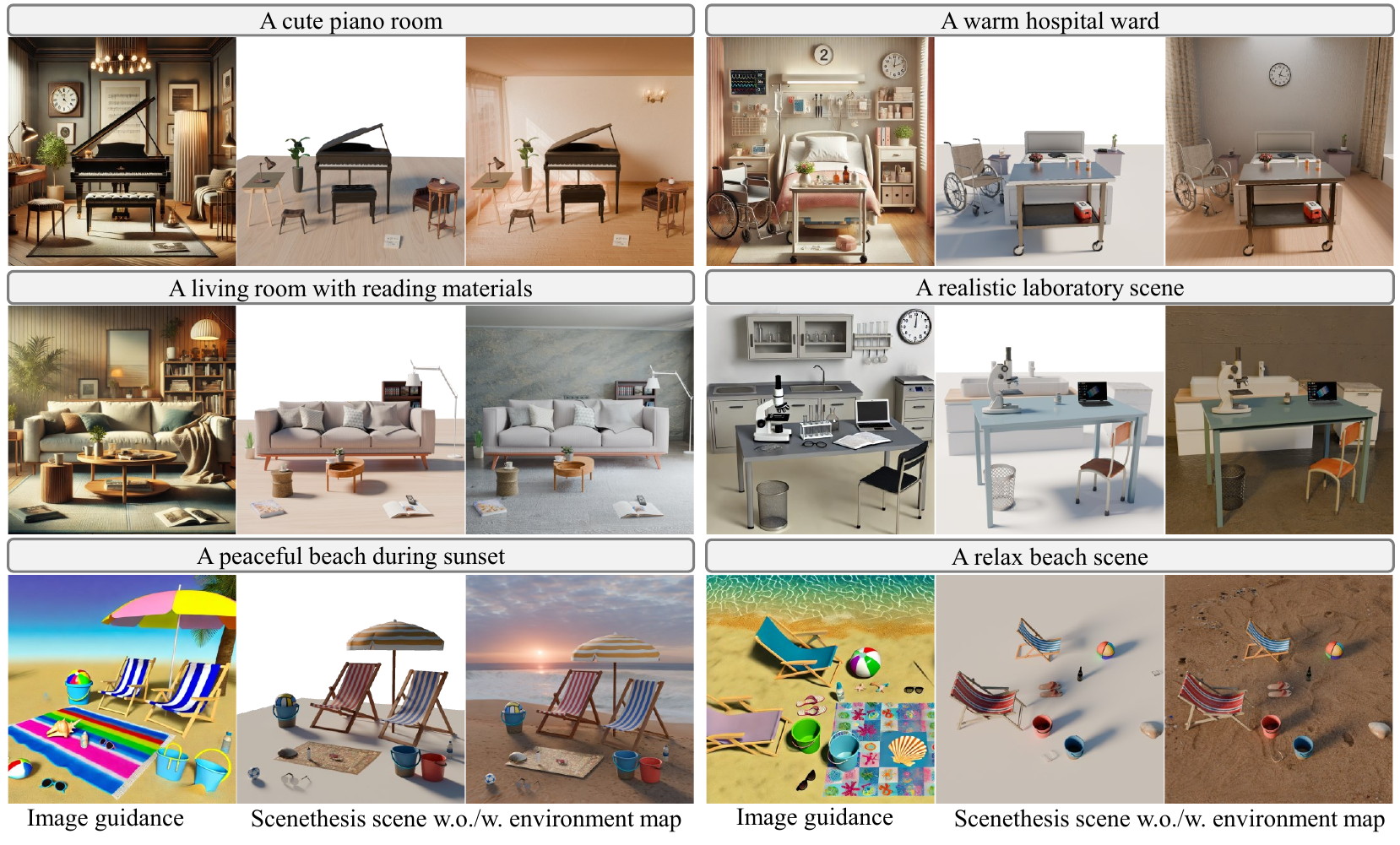}
    \caption{We provide a visual illustration of the generated scenes and their corresponding image guidance. The first column displays the image guidance, while the second and third columns show the generated scenes without and with the environment map, respectively. Note that \name\ focus on layout planning for objects on the ground. Additionally, certain unique assets, such as a beach mat, are unavailable in Objaverse~\cite{deitke2023objaverse}, which may result in the generated scene differing from the image guidance. Future work could enhance the system by incorporating a wider range of assets.}
    \label{fig:image_guidance_genraated_scene}
\end{figure*}

\begin{itemize}
    \item \textbf{Qualitative Results of \name's Scene:} We present different camera views to showcase the qualitative results of \name's scenes, as shown in~\autoref{fig:qualitative_mutliview}. ~\autoref{fig:image_guidance_genraated_scene} presents the generated scenes by \name\ with their image guidance. Note that \name\ focuses on layout planning for ground objects. The absence of certain unique assets in Objaverse~\cite{deitke2023objaverse} may cause discrepancies between the generated scene and the image guidance. Future work could address this by incorporating more diverse assets.
    \item \textbf{Quantitative Results of Physical plausibility Comparison}: The physical Plausibility quantitative comparison presented in Table 1 of the \textbf{Experiment} section. While Holodeck applies both soft and hard constrains based on the Depth-First-Search Solver and small objects are placed on predefined locations. These small objects may collide with each other due to the shape and size variations as shown in~\autoref{fig:holodeck_collide}. 
    \item \textbf{Visual Comparison with Holodeck:} In addition to the quantitative comparison presented in Table 1 of the \textbf{Experiment} section, we provide a visual comparison between \name\ and Holodeck, a state-of-the-art LLM-based 3D interactive scene generation method, in ~\autoref{fig:holodeck_comp}. Based on the four evaluation metrics detailed in the \textbf{Experiment} section, scenes generated by \name\ demonstrate greater diversity in object categories, quantities, and sizes. More importantly, \name's scenes have a broader range of spatial relationships, such as \textit{``on top of''}, \textit{``inside''}, and \textit{``under''}, compared to those generated by Holodeck~\cite{yang2024holodeck}, which supports only \textit{``on top of''} spatial relation.  Furthermore, \name's scenes align more faithfully with the intended scene type. \ie when given the description \textit{``a peaceful beach during sunset''}, \name\ produces an outdoor scene with appropriate beach elements, while Holodeck incorporates beach-related objects but generates an environment resembling an indoor setting.
\end{itemize}

\clearpage

\begin{figure*}
\section{Prompts Examples}
\subsection{Coarse Scene Planning Instruction Prompts}
\begin{tcolorbox}[mybox, title=Coarse Scene Planning Instruction Prompts, label=box:coarse-instruction ]
\textbf{Task Description:}\\
You are responsible for generating a set of common objects and planning a scene based on these common objects. You will be given a list that includes all available object categories and a text prompt to describe a scene. This is a hard task, please think deeply and write down your analysis in following steps:
\begin{enumerate}[leftmargin=3.5em, label=\textbf{Step \arabic*:}]
    \item \textbf{Review All Categories}
    \begin{enumerate}[label=\alph*.]
        \item Begin by thoroughly reviewing the categories in the provided list.
        \item Identify potential groups or clusters of objects within this list that are commonly found in similar environments (e.g., furniture, electronics, household items, etc.).
    \end{enumerate}
    
    \item \textbf{Interpret Input Prompt}
    \begin{enumerate}[label=\alph*.]
        \item Carefully read the input prompt. Understand the theme, primary activities, or the setting it describes, as these will guide your object selection. \ie if the prompt gives: \textit{children playing room}, then you may think of objects like tent, toy, bear, ball, chair, etc.
    \end{enumerate}
    
    \item \textbf{Object Selection}
    \begin{enumerate}[label=\alph*.]
        \item Based on the description, select at least 15 object categories from the list that match the scene.
        \item Determine the anchor object:
        \begin{enumerate}[label=\roman*.]
            \item Identify the anchor object among the selected objects. Consider the following factors:
            \begin{enumerate}[label=\arabic*.]
                \item A large object directly on the ground (\ie floor, table, or shelf).
                \item An object that influences where other objects are placed (\ie a table in a dining room, and there are cups and fruits on the table).
                \item The object should logically anchor the scene and often defines the scene's layout orientation. \ie the sofa in a front-facing view in the scene.
            \end{enumerate}
        \end{enumerate}
    \end{enumerate}

    \item \textbf{Object Cross-check}
    \begin{enumerate}[label=\alph*.]
        \item I will give you \$100 tips if you can cross-check whether objects in the scene can be found in the given category list or its relevant categories. \ie, if there is a bookshelf in your planned scene, the bookshelf should also be found in the given list, or bookcase can be found in the list if bookshelf is not covered by the category. Otherwise, re-plan the scene.
    \end{enumerate}
    
    \item \textbf{Plan Scene with Selected Objects}
    \begin{enumerate}[label=\alph*.]
        \item Based on the description and selected objects, plan the scene, keeping these aspects in mind:
        \begin{enumerate}[label=\roman*.]
            \item \textbf{Functionality}: Choose objects that are contextually relevant to the scene (e.g., selecting a table, chair, flower vase, and utensils for a dining room), but do not generate any wall décor objects.
            \item \textbf{Spatial Hierarchy}: 
            \begin{enumerate}[label=\arabic*.]
                \item Please have a depth effect in the layout. For the depth effect, the scene should have some objects placed on the ground as the background, central, and in the front, resulting in a depth layout. \ie the sofa and bookshelf are the background of the table and chair set in the living room.
                \item Please have a supportive item in the layout. \ie the shoes, bag, and hat are in the display shelf in a clothes store, where the display shelf is a supportive item.
            \end{enumerate}
            \item \textbf{Balance}: Ensure a mix of large and small objects to avoid overcrowding or under-populating the scene. \ie taking the table as the center, there are flower vases, fruits, and cups on the table, and chairs are on the sides.
        \end{enumerate}
    \end{enumerate}
    
    \item \textbf{Output Format:} 
    \begin{enumerate}[label=\alph*.]
        \item Save the selected objects as a json file follow the output format:
        \begin{quote}
        \textbf{Anchor object}:  \\
        \textbf{Other common objects}: 
        \end{quote}
    \item Save scene planning as txt file.
    \end{enumerate}
\end{enumerate}

\end{tcolorbox}
\end{figure*}
\clearpage

\begin{figure*}
\subsection{Coarse Scene Planning Output Example}
\begin{tcolorbox}[mybox, title=Coarse Scene Planning Output Example, label=box:coarse-scene-planning-output-example]
\textbf{User input: a gym room}\\
\begin{enumerate}[leftmargin=3.5em, label=\textbf{Step \arabic*:}]
\item  \textbf{Reviewing Object Categories for Gym-Related Items} 
\newline
From the provided list, here are the relevant categories typically found in a gym setting:
\begin{itemize}
    \item Exercise equipment: dumbbell, treadmill, weight, barbell
    \item Furniture or storage items: bench, shelf, cabinet
    \item Accessories and smaller items: towel, water bottle, yoga mat, scale, bag
\end{itemize}
\item  \textbf{Scene Interpretation} 
\newline
The scene is a gym room, so the primary objects would focus on equipment for strength and cardio exercises, along with supportive items like shelves for organization and accessories like towels and water bottles for hydration.

\item  \textbf{Scene Layout and Object Selection} 
\begin{itemize}
    \item Anchor object: Treadmill (as the main focus of cardio equipment)
    \item Other common objects: Dumbbell, Barbell,Bench, Shelf, Water bottle, Towel, Weight, Bag, Mat, Scale, Cabinet, Exercise ball, Fan, Mirror
\end{itemize}

\item  \textbf{Object cross-check} 
\begin{itemize}
    \item Treadmill - \GreenCheck Available as "treadmill"
    \item Dumbbell - \GreenCheck Available as "dumbbell"
    \item Barbell – \GreenCheck Available as "barbell"
    \item Bench - \GreenCheck Available as "bench"
    \item Shelf - \GreenCheck Available as "shelf"
    \item Water bottle – \GreenCheck Available as "water bottle"
    \item Towel – \GreenCheck Available as "towel rack" 
    \item Weight – \GreenCheck Available as "weight"
    \item Bag – \GreenCheck Available as "bag"
    \item Mat – Not explicitly listed, but a close alternative like "place mat" 
    \item Scale – \GreenCheck Available as "scale (measuring instrument)"
    \item Cabinet – \GreenCheck Available as "cabinet"
    \item Exercise ball – Not explicitly listed, but a close alternative like by "ball"
    \item Fan – \GreenCheck Available as "fan"
    \item Mirror – \GreenCheck Available as "mirror"
\end{itemize}
\item  \textbf{Plan Scene with Selected Objects} 
\begin{itemize}
    \item Background: Shelf and cabinet are placed to the side or back corner of the room for organization, holding smaller items. Mirror mounted on the back wall for checking exercise form. They serves as supportive equipments...It enhances depth and makes the room feel more spacious...
    \item Middle Ground: Treadmill is the primary cardio equipment and anchor object, anchoring the scene. Other objects such as bench and barbell are placed around the treadmill, allowing access from multiple directions for various exercises. They provide visual weight to the scene and represent strength-focused equipment...
    \item Foreground: Dumbbell, water bottle and other objects set closer to the viewer, easily accessible for hand-held exercises. It introduces a smaller, more detailed object in the front. They placed nearby, adding another layer of exercise options for core stability...
\end{itemize}
This hierarchy ensures a balanced layout, where larger and bulkier items form the background... The middle ground includes the treadmill and training equipment, while smaller and accessible items like dumbbells occupy the foreground... This setup provides a clear layout for practical use and visual depth in the gym scene...
\item  \textbf{Output Format}
A json file that include the available objects and a txt file that describe the scene planning.
\end{enumerate}
\end{tcolorbox}
\end{figure*}

\clearpage

\begin{figure*}
\subsection{Instruction for Evaluating Generated Scenes}
\begin{tcolorbox}[mybox, title=GPT-4o Evaluation Prompts for Generated Scenes, label=box:scene-evaluation]
\textbf{Evaluate Generated Scenes for \name\ and Holodeck}
\newline
This is a hard problem. You are supposed to compare the alignment of a pair of images with a given text prompt that describe the scene.  Images contain generated scenes by two different methods. Please evaluate them in the following five aspects:
\begin{enumerate}
    \item \textbf{Object diversity}: Counting number of object and object types in the scene. The higher number, the better object diversity. \ie if there are three shelves and a box in the scene. Then the object type is 2 and the number of counting is 3+1 = 4.
    \item \textbf{Layout coherence}: whether the objects position and orientation in the scene are realistic and adhere to common sense.
    \begin{itemize}
        \item An ideal layout would be dependents on the scene type. \ie the objects in the garage scene will forms as a bit chaotic organization and it would be less reasonable if all objects standing against the wall in a very clean order.
        \item Objects should be placed reasonably. \ie shelves fly in the air or hange on the wall.
    \end{itemize}
    \item \textbf{ Spatial realism \& complexity}: whether scene contains diverse hierarchy. The measure of spatial complexity is by review scenes and carefully evaluate objects relations. The higher diverse relation indicate a better spatial complexity. The spatial hierarchy refers to the following aspects:
    \begin{itemize}
        \item relations such as on the top of, in side of, under the, etc.
        \item objects in various sizes. \ie the scene should contain large objects such as shelf and small objects such as apple.
        \item objects that are arranged in different area and form a depth effects. \ie objects are in the background, in the front, and in the center of the space.
    \end{itemize}
    \item \textbf{Overall performance}: assets' categorizes and styles are faithful to the scene type. \ie fruits are less likely in the warehouse, while carts, boxes are more likely in a warehouse scene.  Thus, it would be a bad performance if find fruits in the warehouse and good performance if carts\/shelf\/boxes presents in the warehouse. Please carefully evaluate the object's category and scene type. To see whether it is reasonable enough based on your common sense.
    \item Please ignore aspects in : environment background style such as lighting, shadow, color, and decorations in the wall etc. This is a challenging task, please think aloud and make your decisions by:
    \begin{itemize}
        \item analyzing each criteria one by one for each image.
        \item report your evaluation results based on the analysis.
    \end{itemize}
    \item Your \textit{output format} should be: \{\} is better than \{\} in the criteria \{\}.
\end{enumerate}
\end{tcolorbox}
\end{figure*}

\clearpage
\begin{figure*}
\subsection{Evaluation Example of Generated Scenes}
\begin{tcolorbox}[mybox, title=GPT-4o Evaluation Results, label=box:scene_evaluation_results]
\textbf{Scenethesis's Scene}
\begin{enumerate}
    \item \textbf{Object diversity}:
    \begin{itemize}
        \item Objects visible: Ladder, barrel, boxes, fire extinguisher, table, tire, crate.
        \item Object types: 8 types. Total count: 13
        \item Higher diversity due to various objects.
    \end{itemize}
    \item  \textbf{Layout coherence}:
    \begin{itemize}
        \item Objects are dispersed naturally, suggesting activity.
        \item Different sizes and a less structured arrangement suggest a more realistic warehouse scene.
    \end{itemize}
    \item \textbf{Spatial realism \& complexity}:
    \begin{itemize}
        \item Objects have various spatial relations; ladder stands, boxes on ground, tables at center...
        \item Good diversity with a mix of background and central objects.
    \end{itemize}
    \item \textbf{Overall performance}:
    \begin{itemize}
        \item Varied objects like ladders, barrels, crates, and extinguishers are typical in a warehouse.
        \item Realistic depiction of a warehouse setting.
    \end{itemize}
\end{enumerate}
\textbf{Holodeck's Scene}
\begin{enumerate}
    \item \textbf{Object diversity}:
    \begin{itemize}
        \item Objects visible: Shelves, cabinets, safety vests, and workbenches.
        \item Object types: Shelves, cabinets, vests, workbenches. Total count: 7+2=9 (shelves are considered per structure, as are others).
        \item Diversity is somewhat low as there are few types.
    \end{itemize}
    \item \textbf{Layout coherence}:
    \begin{itemize}
        \item Organization is neat; workbenches and cabinets are along the walls.
        \item Reasonable layout for a tidy scene, but lacks variety in sizes.
    \end{itemize}
    \item \textbf{Spatial realism \& complexity}:
    \begin{itemize}
        \item Shelving and cabinets against walls create vertical space; workbenches on the ground.
        \item Lacks diverse spatial relations.
    \end{itemize}
    \item \textbf{Overall performance}: Objects like shelves and workbenches fit a warehouse setting, but it lacks variety typical in larger-scale warehousing.
\end{enumerate}
\textbf{Evaluation}   
\newline
\textbf{Object diversity}: \name\ is better than Holodeck; \textbf{Layout coherent}: \name\ is better than Holodeck; \textbf{Spatial realism \& complexity}: \name\ is better than Holodeck; \textbf{Overall performance}: \name\ is better than Holodeck.

\end{tcolorbox}
\end{figure*}
\clearpage

\begin{figure*}
\subsection{Instruction Prompts for Ablation Study}
\begin{tcolorbox}[mybox, title=Pose Alignment Evaluation Instruction Prompt, label=box:spatial_alignment] This task involves evaluating the pose alignment between two images in a pair. One image serves as the image guidance (GT), while the other is a generated image. Your objective is to measure the pose alignment of the generated image relative to the GT image. Follow these steps for evaluation:
\begin{enumerate}
    \item \textbf{Review Objects in the GT Image}: Examine all objects in the GT image, focusing on their locations, sizes, and orientations. Understand the spatial relationships among objects, such as \textit{on top of}, \textit{inside}, \textit{under}, etc.
    \item \textbf{Evaluate pose alignment}: Assess the similarity between the generated image and the GT image based on the following three aspects:
    \begin{itemize}
        \item Location and Size Similarity: Compare the location and size of objects in the generated image with those in the GT image. Assign a similarity score between 0 and 1, where 1 indicates the highest similarity. For example:
        \begin{itemize}
            \item If an apple in the GT image is placed at the center of a table, and in the generated image it is placed on the left side of the table, the similarity might be moderate (e.g., 0.5).
            \item If the apple is misplaced (e.g., on the ground or missing entirely), the similarity would be very low (e.g., 0.1).
        \end{itemize}
        \item Orientation Similarity: Examine the orientation of each object in the generated image compared to the GT image. Pay close attention to details, noting any deviations such as slight tilts (e.g., right/left, up/down) or rotations that create different perspectives. Assign a score from 0 to 1, where 1 indicates perfect alignment and 0 indicates a significant mismatch (e.g., opposite orientation).
        \item Overall Layout Similarity: Assess the overall visual coherence of the generated image compared to the GT image, including spatial relationships and hierarchical structure. Assign a similarity score between 0 and 1, where 1 represents a perfect match. For instance:
        \begin{itemize}
            \item A perfect match occurs when the generated image maintains the same spatial relationships, relative locations, sizes, and orientations as the GT image (e.g., an apple placed at the center of a table in both images).
            \item Small deviations in placement or orientation are acceptable but should result in a lower score.
        \end{itemize}
    \end{itemize}
    \item \textbf{Exclusions}: Do not consider style, appearance, object shape, or texture in your evaluation. Focus solely on pose alignment.
    \item \textbf{Output Format}: Clearly document your similarity scores for each aspect (Location and Size Similarity, Orientation Similarity, and Overall Layout Similarity) following the format: location and size similarity score is \{\}, orientation similarity score is \{\}, and overall layout similarity score is \{\}. Please save the evaluated scores as a json file.
\end{enumerate}
\end{tcolorbox}
\end{figure*}
\clearpage

{
    \small
    \bibliographystyle{ieeenat_fullname}
    \bibliography{main}
}

\end{document}